\documentclass[sigconf]{acmart}

\usepackage{balance}

\usepackage{amssymb}
\usepackage{amsmath}

\usepackage{booktabs}
\usepackage{multirow}
\usepackage{subfigure}
\usepackage{cleveref}
\crefname{table}{Table}{Tables}
\crefname{algorithm}{Algorithm}{Algorithms}
\crefname{figure}{Figure}{Figures}
\setcopyright{acmcopyright}
\usepackage{balance}

\def\BibTeX{{\rm B\kern-.05em{\sc i\kern-.025em b}\kern-.08emT\kern-.1667em\lower.7ex\hbox{E}\kern-.125emX}}
    
\begin{document}

\fancyhead{}

\title{Heterogeneous Domain Adaptation via Soft Transfer Network}


\author{Yuan Yao}
\affiliation{ 
     School of Computer Science and Technology, Harbin Institute of Technology, Shenzhen, China
 }
 \email{yaoyuan@stu.hit.edu.cn}

 \author{Yu Zhang}
\affiliation{ 
     Department of Computer Science and Engineering, Southern University of Science and Technology, Shenzhen, China
 }
 \email{yu.zhang.ust@gmail.com}

 \author{Xutao Li}
\affiliation{ 
     School of Computer Science and Technology, Harbin Institute of Technology, Shenzhen, China
 }
 \email{lixutao@hit.edu.cn}

 \author{Yunming Ye}
 \authornote{Corresponding Author.}
\affiliation{ 
     School of Computer Science and Technology, Harbin Institute of Technology, Shenzhen, China
 }
 \email{yeyunming@hit.edu.cn}

%
\renewcommand{\shortauthors}{Trovato and Tobin, et al.}

%
\begin{abstract}
Heterogeneous domain adaptation (HDA) aims to facilitate the learning task in a target domain by borrowing knowledge from a heterogeneous source domain. In this paper, we propose a Soft Transfer Network (STN), which jointly learns a domain-shared classifier and a domain-invariant subspace in an end-to-end manner, for addressing the HDA problem. The proposed STN not only aligns the discriminative directions of domains but also matches both the marginal and conditional distributions across domains. To circumvent negative transfer, STN aligns the conditional distributions by using the soft-label strategy of unlabeled target data, which prevents the hard assignment of each unlabeled target data to only one category that may be incorrect. Further, STN introduces an adaptive coefficient to gradually increase the importance of the soft-labels since they will become more and more accurate as the number of iterations increases. We perform experiments on the transfer tasks of image-to-image, text-to-image, and text-to-text. Experimental results testify that the STN significantly outperforms several state-of-the-art approaches.
\end{abstract}

%
%

\ccsdesc[500]{Computing methodologies~Transfer learning}
\ccsdesc[300]{Computing methodologies~Object recognition}
\ccsdesc[100]{Computing methodologies~Semi-supervised learning settings}

\copyrightyear{2019} 
\acmYear{2019} 
\acmConference[MM '19]{Proceedings of the 27th ACM International Conference on Multimedia}{October 21--25, 2019}{Nice, France}
\acmBooktitle{Proceedings of the 27th ACM International Conference on Multimedia (MM '19), October 21--25, 2019, Nice, France}
\acmPrice{15.00}
\acmDOI{10.1145/3343031.3350955}
\acmISBN{978-1-4503-6889-6/19/10}

%
\keywords{Heterogeneous domain adaptation, soft-label, adaptive coefficient, subspace learning}

%

%
\maketitle


\section{Introduction}

\begin{figure}[t]
  \setlength{\abovecaptionskip}{2pt} 
  \centering
  \includegraphics[width=3.1in]{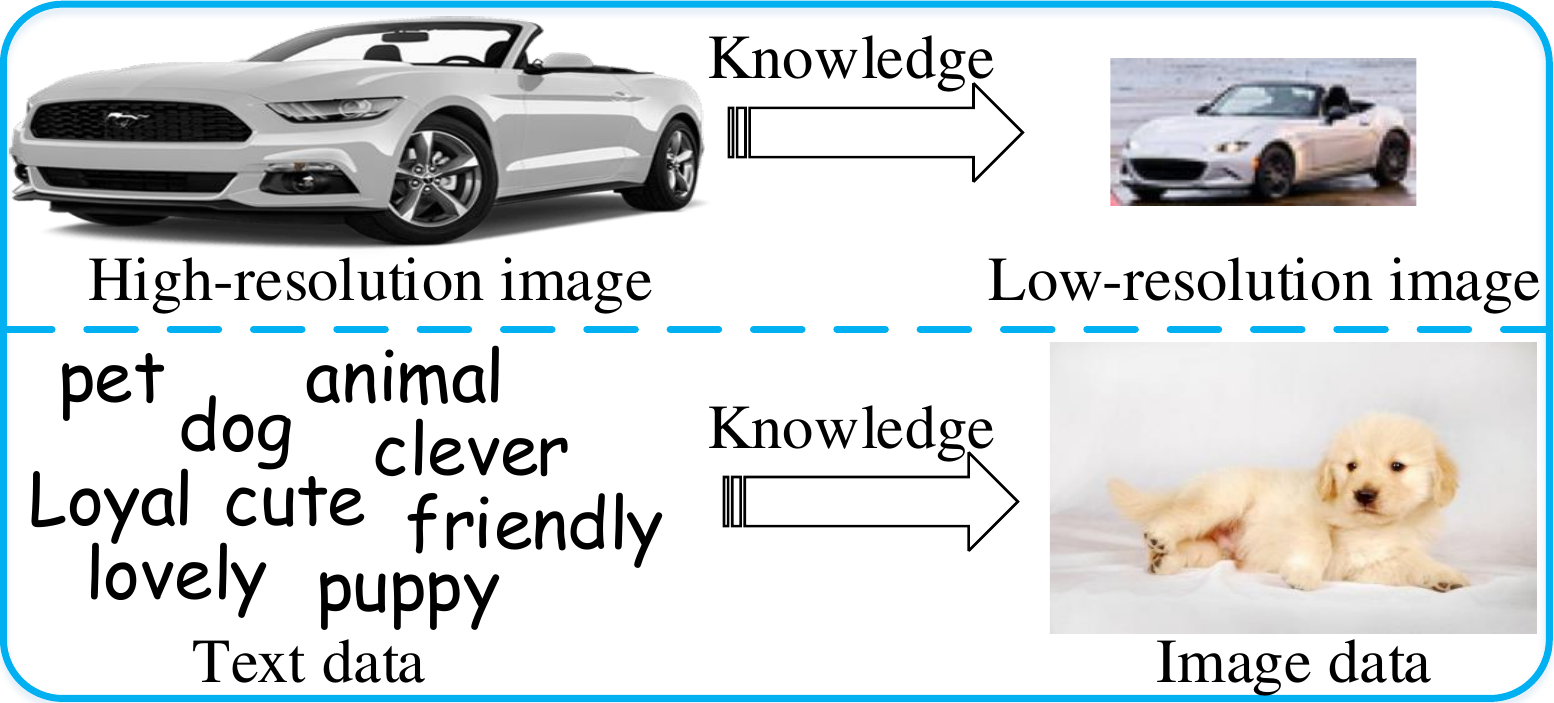}
  \caption{We aim to transfer knowledge between different feature spaces. Top row: transferring the knowledge from high-resolution images to low-resolution ones. Bottom row: transferring the knowledge from text data to image data.}
  \label{fig:exampleImage}
  \vspace{-4.mm}
  \end{figure}

The success of supervised learning mostly depends on the availability of sufficient labeled data. However, in many real-world applications, it is often prohibitive and time-consuming to acquire enough labeled data. To alleviate this concern, \emph{domain adaptation} (DA) models \cite{Pan-2010,Weiss-2016,Csurka-2017,Day-2017} are proposed to facilitate the learning task in a label-scarce domain, \emph{i.e.}, the target domain, by borrowing knowledge from a label-rich and related domain, \emph{i.e.}, the source domain. They have achieved a series of successes in fields such as visual recognition \cite{Zhu-2011,Tan-2015,Tan-2017,Long-2013,Long-2015,Long-2017,Wang-2018,Ding-2018} and text categorization \cite{Xiao-2015,Zhou-2014}. As a rule, most existing studies \cite{Long-2013,Long-2014,Long-2015,Long-2017,Long-2018,Bousmalis-2016,Cao-2018,Ding-2018,Wang-2018} assume that the source and target domains share the same feature space. In reality, however, it is often not easy to seek for a source domain with the same feature space as the target domain of interest. Hence, we focus in this paper on a more general and challenging scenario where the source and target domains are drawn from different feature spaces, which is known as \emph{heterogeneous domain adaptation} (HDA). For example, the source and target images are characterized by diverse resolutions (\textit{e.g.}, the top row in \cref{fig:exampleImage}); similarly, the source domain is textual whereas the target one is visual (\textit{e.g.}, the bottom row in \cref{fig:exampleImage}).

To bridge between two heterogeneous domains, existing HDA works typically choose to project data from one domain to another \cite{Xiao-2015,Kulis-2011,Hoffman2013,Hoffman-2014,Zhou-2014,Tsai-2016a,Tsai-2016}, or find a domain-invariant subspace \cite{Yao-2015,Xiao-2015a,Shi-2010,Wang-2011,Duan-2012,Li-2014,Hsieh-2016}. On one hand, projections are learned in \cite{Duan-2012,Li-2014,Hoffman2013,Hoffman-2014} by performing the classifier adaptation strategy (\emph{i.e.}, training a domain-shared classifier with labeled cross-domain data). However, the type of methods does not explicitly minimize the distributional discrepancy across domains. On the other hand, another line of studies \cite{Wang-2011,Tsai-2016a} only considers the distribution matching strategy (\emph{i.e.}, reducing the distributional divergency between domains), but ignores the classifier adaptation one. Thus, this kind of methods does not directly align the discriminative directions of domains. Moreover, several works \cite{Hsieh-2016,Tsai-2016} derive the projections by iteratively performing the classifier adaptation and distribution matching. However, their performance is not stable since the iterative combination is a bit heuristic. In addition, they use trained classifier to assign each unlabeled target data to only one class that may be incorrect during the distribution matching, which is risky to negative transfer. Although Xiao and Guo \cite{Xiao-2015a} jointly consider the classifier adaptation and distribution matching in a unified framework, it does not further reduce the divergence between conditional distributions, which may be more important to the classification performance than aligning the marginal distributions. Note that all the above methods are shallow learning models that cannot learn more powerful feature representations. Although many deep learning models \cite{Long-2015,Long-2017,Long-2018,Bousmalis-2016,Ding-2018} are developed for domain adaptation, they cannot be directly applied to solve the HDA problem. Only few works \cite{Shu-2015,Chen-2016} solve the HDA problem by utilizing deep learning models. However, \cite{Shu-2015} requires the co-occurrence cross-domain data and \cite{Chen-2016} does not consider minimizing the distributional divergency across domains. 

  \begin{figure}[t]
  \setlength{\abovecaptionskip}{2pt} 
  \centering
  \includegraphics[width=3.2in]{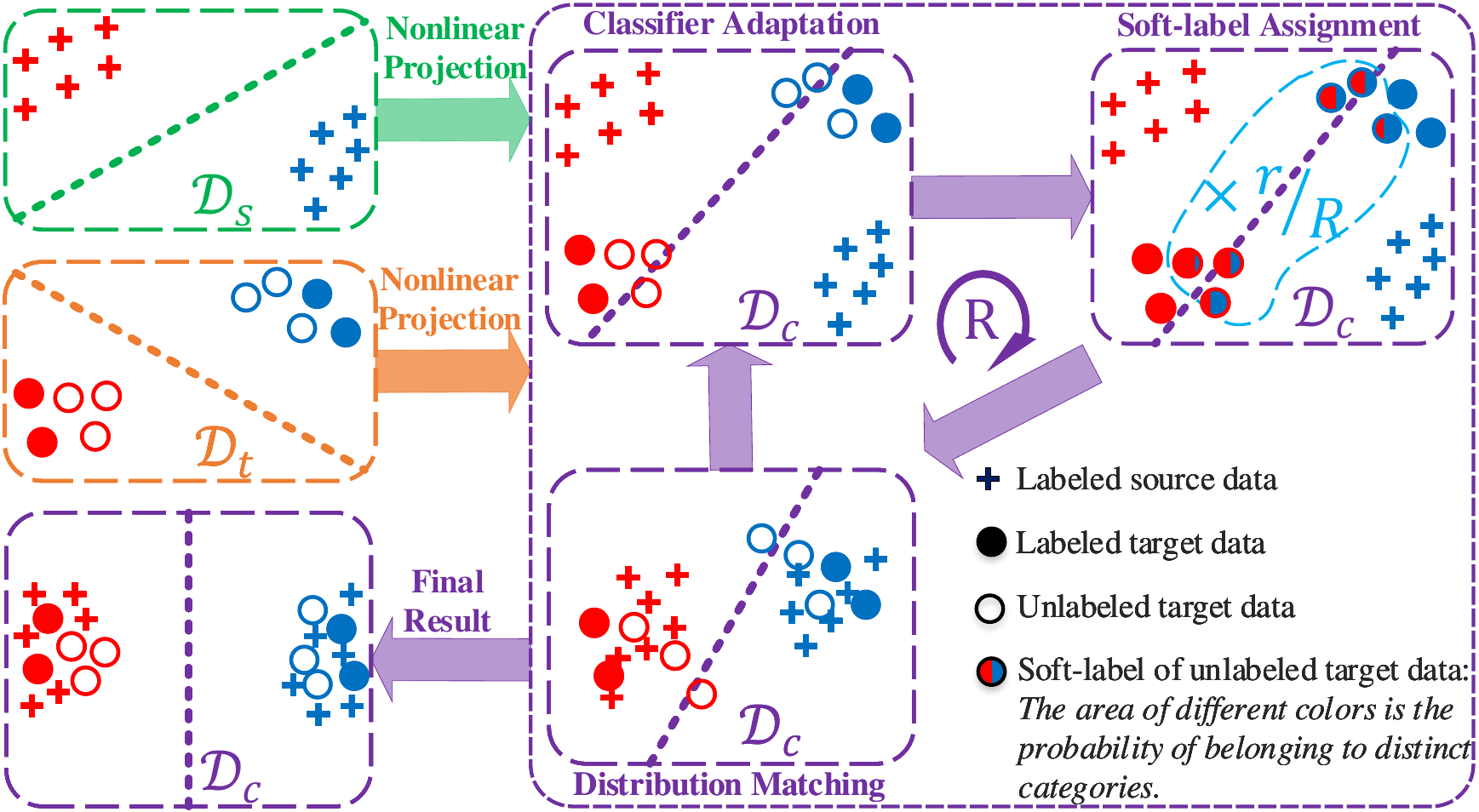}
  \caption{An illustrative example of STN. 
  Here, $\mathcal{D}_s$, $\mathcal{D}_t$, and $\mathcal{D}_c$ denote the source domain (in green), the target domain (in orange), and the common subspace (in purple), respectively. $\left. r \middle/ R \right.$ is the coefficient of the soft-label strategy to adaptively increase its importance, where $R$ is the total number of iterations and $r$ is the index of current iteration. Three key strategies are designed in STN for learning a domain-invariant $\mathcal{D}_c$. The classifier adaptation strategy aligns the discriminative directions of domains. The soft-label strategy avoids the hard assignment of each unlabeled target data to only one class that may be incorrect. The distribution matching strategy reduces the distributional divergence.}
  \label{fig:rationale}
  \vspace{-4.5mm}
  \end{figure}

To overcome the limitations of the above methods, we propose a \emph{Soft Transfer Network} (STN) for solving the HDA problem. The core rationale is depicted in \cref{fig:rationale}. In STN, we first project both the source and target data into a common subspace by using multiple nonlinear transformations. Then, a domain-invariant subspace is learned by jointly performing three key strategies: classifier adaptation, soft-label assignment, and distribution matching. The first stategy trains a domain-shared classifier with projected labeled source and target data by aligning the discriminative directions of domains. The second one assigns the \emph{soft-label} (\emph{i.e.}, a probability distribution over all the categories) to each unlabeled target data, which prevents the hard assignment of each unlabeled target data to only one class that may be incorrect. In addition, we observe that the soft-labels will be more and more accurate as more iterations are executed. Thus, an adaptive coefficient is used to gradually increase the importance of the soft-labels. The last one matches both the marginal and conditional distributions between domains. Finally, we obtain a domain-shared classifier and a domain-invariant subspace for categorizing unlabeled target data. 

The major contributions of this paper are three-fold. \textbf{(1)} To the best of our knowledge, there is no existing work that utilizes the \emph{soft-label} strategy of unlabeled target data to address the HDA problem. \textbf{(2)} We propose a STN to learn a domain-shared classifier and a domain-invariant subspace in an end-to-end manner. \textbf{(3)} Extensive experimental results are presented on the transfer tasks of image-to-image, text-to-image, and text-to-text to verify the effectiveness of the proposed STN.

\section{Related Work} \label{RelatedWork}

As aforementioned, existing HDA models either project data from one domain to the other or derive a domain-invariant subspace. Accordingly, they can be grouped into two categories, namely asymmetric transformation method and symmetric transformation method.

\noindent \textbf{Asymmetric transformation method}. Kulis \emph{et al.} \cite{Kulis-2011} present an Asymmetric Regularized Cross-domain Transformation (ARC-t) to project data from one domain to the other by learning an asymmetric nonlinear transformation. Zhou \emph{et al.} \cite{Zhou-2014} introduce a Sparse Heterogeneous Feature Representation (SHFR) to project the weights of classifiers in the source domain into the target one via a sparse transformation. Hoffman \emph{et al.} \cite{Hoffman2013,Hoffman-2014} propose a Max-Margin Domain Transform (MMDT), which maps the target data into the source domain by training a domain-shared support vector machine. Similarly, Xiao and Guo \cite{Xiao-2015} develop a Semi-Supervised Kernel Matching for Domain Adaptation (SSKMDA) to transform the target data to similar source data, where a domain-shared classifier is simultaneously trained. Tsai \emph{et al.} \cite{Tsai-2016a} propose a Label and Structure-consistent Unilateral Projection (LS-UP) model. This method maps the source data into the target domain with the aim of reducing the distributional difference and preserving the structure of projected data. They also propose a Cross-Domain Landmark Selection (CDLS) \cite{Tsai-2016}, where the representative source and target data are identified while reducing the distributional divergence between domains. Very recently, a few HDA approaches are presented based on the theory of Optimal Transport (OT) \cite{Villani-2008}. For example, Ye \emph{et al.} \cite{Ye-2018} develop a Metric Transporation on HEterogeneous REpresentations (MAPHERE) approach, which jointly learns an asymmetric transformation and an optimal transportation. Yan \emph{et al.} \cite{Yan-2018} propose a Semi-supervised entropic Gromov-Wasserstein (SGW) discrepancy to transport source data into the target domain. Although this method utilizes the supervision information to guide the learning of optimal transport, it does not consider unlabeled target data when matching the conditional distributions between domains.  

\noindent \textbf{Symmetric transformation method}. Shi \emph{et al.} \cite{Shi-2010} develop a Heterogeneous spectral MAPping (HeMAP) to learn a pair of projections based on spectral embeddings. Wang and Mahadevan \cite{Wang-2011} propose a Domain Adaptation using Manifold Alignment (DAMA) to find projections by preserving both the topology of each domain and the discriminative structure. Duan \emph{et al.} \cite{Duan-2012} put forward a Heterogeneous Feature Augmentation (HFA) approach. This method works by first augmenting the projected data with the original data features and then training a domain-shared support vector machine with the augmented features. Subsequently, HFA is generalized to a semi-supervised extension named SHFA \cite{Li-2014} for effectively leveraging unlabeled target data. Xiao and Guo \cite{Xiao-2015a} present a Subspace Co-Projection with ECOC (SCP-ECOC) for HDA to learn projections by both training a domain-shared classifier and reducing the discrepancy on marginal distributions. Yao \emph{et al.} \cite{Yao-2015} develop a Semi-supervised Domain Adaptation with Subspace Learning (SDASL) to learn projections by simultaneously minimizing the classification error, preserving the locality information of the original data, and considering the manifold structure of the target domain. Hsieh \emph{et al.} \cite{Hsieh-2016} introduce a Generalized Joint Distribution Adaptation (G-JDA) model. This method reduces the difference in both the marginal and conditional distributions between domains while learning projections. Yan \emph{et al.} \cite{Yan-2017} present a Discriminative Correlation Analysis (DCA) to jointly learn a discriminative correlation subspace and a target domain classifier. Recently, Li \emph{et al.} \cite{Li-2018} put forward a Progressive Alignment (PA) approach. This method first learns a common subspace by dictionary-shared sparse coding, and then mitigates the distributional divergence between domains. Besides, some HDA methods are developed based on deep learning models. For instance, Shu \emph{et al.} \cite{Shu-2015} propose a weakly-shared Deep Transfer Network (DTN) to address the learning of heterogeneous cross-domain data, but it requires the co-occurrence cross-domain data. Recently, Chen \emph{et al.} \cite{Chen-2016} put forward a Transfer Neural Trees (TNT) for HDA, which jointly addresses cross-domain feature projection, adaptation, and recognition in an end-to-end network. However, TNT does not explicitly minimize the distributional divergence across domains. 

Moreover, our work is related to a few homogeneous DA studies that utilize the pseudo-label strategy of unlabeled target data. Specifically, Long \emph{et al.} \cite{Long-2013} propose a Joint Distribution Adaptation (JDA) to align both the marginal and conditional distributions between domains. As a follow-up, in \cite{Long-2014} they further develop a Adaptation Regularization based Transfer Learning (ARTL) framework. This method works by simultaneously minimizing the structural risk, reducing the distributional divergence, and acquiring the manifold information. Yan \emph{et al.} \cite{Yan-2017a} introduce a weighted MMD to mitigate the effect of class weight bias, and further propose a Weighted Domain Adaptation Network (WDAN) to address the learning of homogeneous cross-domain data. Zhang \emph{et al.} \cite{Zhang-2017} present a Joint Geometrical and Statistical Alignment (JGSA) to reduce both the distributional and geometrical divergence across domains. Wang \emph{et al.} \cite{Wang-2018} propose a Manifold Embedded Distribution Alignment (MEDA) approach, which learns a domain-invariant classifier by both minimizing the structural risk and performing dynamic distribution alignment. Ding \emph{et al.} \cite{Ding-2018a} put forward a Graph Adaptive Knowledge Transfer (GAKT) to jointly learn target labels and domain-invariant features. However, these methods either learn a feature transformation shared by cross-domain data \cite{Long-2013,Long-2014,Yan-2017a,Wang-2018}, or learn two feature transformations with the same size to constrain the divergence between transformations \cite{Zhang-2017,Ding-2018a}, which thus cannot directly deal with the HDA problem.

Overall, our work belongs to the symmetric transformation method. Inspired by the above-mentioned studies, we propose to jointly learn a domain-shared classifier and a domain-invariant subspace in an end-to-end manner, which not only aligns the discriminative directions of domains but also reduces the distributional divergence between domains.

  \begin{figure}[t]
  \setlength{\abovecaptionskip}{2pt} 
  \centering
  \includegraphics[width=3.2in]{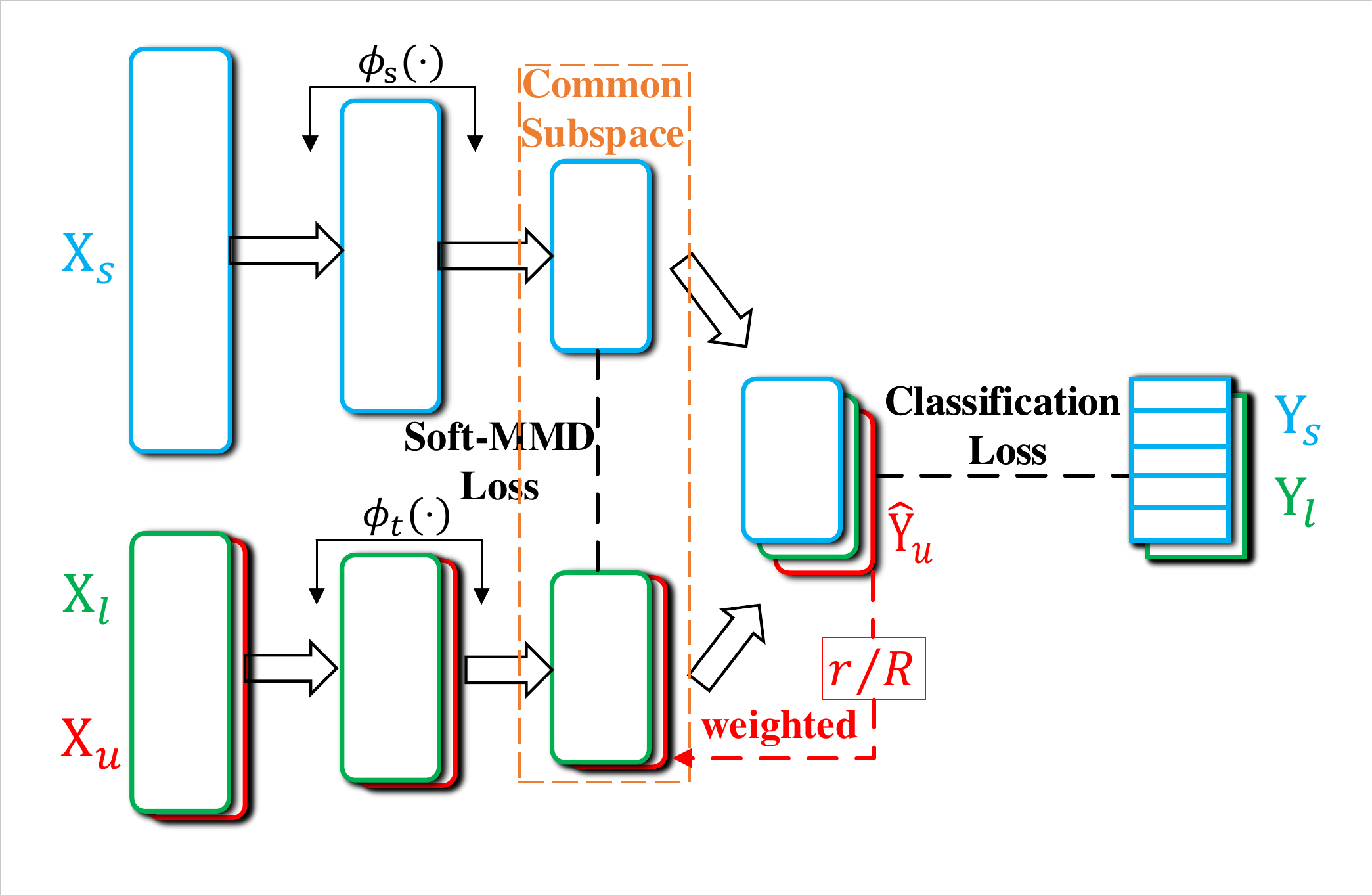}
  \caption{The architecture of the proposed STN, where $\{\mathbf{X}_s, \mathbf{Y}_s\}$ (in blue) and $\{\mathbf{X}_l, \mathbf{Y}_l\}$ (in green) are labeled source and target data, respectively, $\mathbf{X}_u$ (in red) is unlabeled target data, $\widehat{\mathbf Y}_u = f(\phi_t(\mathbf{X}_u))$ (in red) is the soft-label of $\mathbf{X}_u$, $\phi_s(\cdot)$ and $\phi_t(\cdot)$ are feature projection networks of the source and target domains, respectively. Our model consists of two losses: a \emph{classification loss} for aligning the discriminative directions of domains, and a \emph{soft-MMD loss} to reduce the distributional divergence across domains.}
  \label{fig:architecture}
  \vspace{-4.5mm}
  \end{figure}

\section{Soft Transfer Network} \label{SoftTransferNetworks}

In this section, we present the proposed STN. We begin with the definitions and terminologies. The source domain is denoted by $\mathcal{D}_s=\{(\mathbf{x}^s_i,y^s_i)\}_{i=1}^{n_s}$, where $\mathbf{x}^s_i \in \mathbb{R}^{d_s}$ is the $i$-th source domain data with $d_s$-dimensional features, and $y^s_i \in \mathcal{Y}=\{1,2,\cdots,C\}$ is its associated class label with $C$ as the number of classes. Similarly, let $\mathcal{D}_t = \mathcal{D}_l \cup \mathcal{D}_u = \{(\mathbf{x}^l_i,y^l_i)\}_{i=1}^{n_l} \cup \{\mathbf{x}^u_i\}_{i=1}^{n_u}$ be the target domain, where $\mathbf{x}^l_i$ $(\mathbf{x}^u_i) \in \mathbb{R}^{d_t}$ is the $i$-th labeled (unlabeled) target domain data with $d_t$-dimensional features, and $y^l_i \in \mathcal{Y}$ is the corresponding class label. Note that in the HDA problem, we have $d_s \not= d_t$, $n_s \gg n_l$, and $n_u \gg n_l$. The goal is to design a heterogeneous transfer network for predicting the labels of unlabeled target domain data in $\mathcal{D}_u$.

For simplicity of presentation, we denote by $\mathbf{X}_s = [\mathbf{x}_1^s, \cdots, \mathbf{x}_{n_s}^s]^\mathrm{T} \in \mathbb{R}^{n_s \times d_s}$ the data matrix in $\mathcal{D}_s$ and by $\mathbf{Y}_s$ = $[\mathbf{g}_{y_1^s}, \cdots, \mathbf{g}_{y_{n_s}^s}]^\mathrm{T} \in \mathbb{R}^{n_s \times C}$ the corresponding label matrix, where $\mathbf{g}_y \in \mathbb{R}^{C}$ denotes a one-hot vector with the $y$-th element being one. Analogously, let $\mathbf{X}_l = [\mathbf{x}_1^l, \cdots, \mathbf{x}_{n_l}^l]^\mathrm{T} \in \mathbb{R}^{n_l \times d_t}$ be the data matrix in $\mathcal{D}_l$ with an associated label matrix $\mathbf{Y}_l = [\mathbf{g}_{y_1^l}, \cdots, \mathbf{g}_{y_{n_l}^l}]^\mathrm{T} \in \mathbb{R}^{n_l \times C}$ and $\mathbf{X}_u = [\mathbf{x}_1^u, \cdots, \mathbf{x}_{n_u}^u]^\mathrm{T} \in \mathbb{R}^{n_u \times d_t}$ be the data matrix in $\mathcal{D}_u$. We also denote by $\mathbf{X}_t = \left[ \mathbf{X}_l;\mathbf{X}_u \right] \in \mathbb{R}^{n_t \times d_t}$ the data matrix in $\mathcal{D}_t$, where $n_t = n_l + n_u$ is the total number of the target domain data. 

The key challenge of HDA arises in that the feature spaces of the source and target domains are different. Thus, we cannot directly transfer knowledge from the source domain to the target one. To approach this problem, we first assume there is a common subspace with the dimensionality $d$, and then build two projection networks $\phi_s(\cdot): \mathbb{R}^{d_s} \rightarrow \mathbb{R}^{d}$ and $\phi_t(\cdot): \mathbb{R}^{d_t} \rightarrow \mathbb{R}^{d}$, which are used to transform source and target data into the subspace, respectively. We aim to find the optimal projection networks $\phi_s(\cdot)$ and $\phi_t(\cdot)$, which can form a domain-invariant subspace and lead to an effective transfer strategy. As depicted in \cref{fig:architecture}, we achieve this by optimizing an objective function with two components: a \emph{classification loss} for aligning the discriminative directions of domains, and a \emph{Soft-MMD loss} to reduce the distributional divergence across domains. The details of each component will be described in the following sections.

\subsection{Classification Loss}

We apply the classification loss to train a domain-shared classifier $f(\cdot)$ with projected labeled source and target data. It can be used to align the discriminative directions of domains. Further, we propose to jointly learn $f(\cdot)$, $\phi_s(\cdot)$, and $\phi_t(\cdot)$ under the \emph{Structural Risk Minimization} (SRM) framework \cite{Vapnik-1998}, and formulate the classification loss as
\begin{equation} \label{loss:classification}
\small
\mathcal{L}_c \big[ \mathbf{Y}_a, f(\mathbf{X}_a) \big]
= \frac{1}{n_s + n_l} \mathcal{L} \big[ \mathbf{Y}_a, f(\mathbf{X}_a) \big]
+ \tau \big( \left\| \phi_s \right\|^2 + \left\| \phi_t \right\|^2 + \left\| f \right\|^2 \big),
\end{equation}
where $\mathbf{X}_a = \big[ \phi_s(\mathbf{X}_s); \phi_t(\mathbf{X}_l) \big] \in \mathbb{R}^{(n_s+n_l) \times d}$ denotes projected labeled data matrix in two domains, $\mathbf{Y}_a = \big[ \mathbf{Y}_s; \mathbf{Y}_l \big] \in \mathbb{R}^{(n_s+n_l) \times C}$ is the associated label matrix, $\mathcal{L}[\cdot, \cdot]$ is the cross-entropy function, $f(\cdot)$ is the softmax function, and $\tau$ is a positive regularization parameter. According to the SRM principle, the domain-shared classifier $f(\cdot)$ can accurately predict unknown target data by minimizing Eq.~(\ref{loss:classification}), given the domain-invariant subspace.

\subsection{Soft-MMD Loss}

As mentioned above, we expect to discover a domain-invariant subspace by enforcing the discriminative structure of the source domain to be consistent with that of the target domain. If there are sufficient labeled data in both domains, we can make it by only minimizing Eq.~(\ref{loss:classification}). However, the labeled target data is very scarce, which is not enough to identify a domain-invariant subspace. To eliminate this problem, we design a \emph{Soft Maximum Mean Discrepancy} (Soft-MMD) loss, which can be applied to match both the marginal and conditional distributions between domains. As a result, the Soft-MMD loss is formulated as
\begin{equation} \label{loss:SMMD}
\small
\mathcal{L}_s \big[ \phi_s(\mathbf{X}_s), \phi_t(\mathbf{X}_t) \big]
= \mathcal{Q}_m \big[ \phi_s(\mathbf{X}_s), \phi_t(\mathbf{X}_t) \big]
+ \mathcal{Q}_c \big[ \phi_s(\mathbf{X}_s), \phi_t(\mathbf{X}_t) \big],
\end{equation}
where $\mathcal{Q}_m[\cdot,\cdot]$ acts as the divergence between marginal distributions of projected cross-domain data, and $\mathcal{Q}_c[\cdot,\cdot]$ stands for that between conditional distributions.

We first detail how to formulate $\mathcal{Q}_m[\cdot,\cdot]$. The empirical \emph{Maximum Mean Discrepancy} (MMD) \cite{Gretton-2007} has been proven to be a powerful tool for measuring the divergency between marginal distributions. Its exact idea is to calculate the distance between the centroids of both the source and target data, and then uses the distance to model the discrepance on marginal distributions. Thus, we define $\mathcal{Q}_m[\cdot,\cdot]$ as
\begin{equation}
\small
\mathcal{Q}_m \big[ \phi_s(\mathbf{X}_s), \phi_t(\mathbf{X}_t) \big]
= \left\| \frac{1}{n_s} \sum_{i=1}^{n_s} \widetilde{\mathbf{x}}_i^s 
  - \frac{1}{n_t} \sum_{i=1}^{n_t} \widetilde{\mathbf{x}}_i^t \right\|^2,
\end{equation}
 where $\widetilde{\mathbf{x}}_i^s$ and $\widetilde{\mathbf{x}}_i^t$ are the $i$-th projected source and target data, respectively.

 Next, we describe how to design $\mathcal{Q}_c[\cdot,\cdot]$. Since directly modelling the conditional distribution (\emph{i.e.}, $\mathcal{P}(Y|X)$) is difficult, we turn to explore the class-conditional distribution (\emph{i.e.}, $\mathcal{P}(X|Y)$) instead \cite{Long-2013,Tsai-2016,Hsieh-2016}, and apply the divergency between class-conditional distributions to approximate that between conditional distributions. Hence, the divergence on conditional distributions can be approximated by calculating the sum of the distance between centroids of both the source and target data in each category. However, a large amount of target data has no label information. To make full use of unlabeled target data, \cite{Long-2013} first proposes to use the \emph{hard-label} strategy of unlabeled target data, which can be performed by applying a base classifier trained on labeled cross-domain data to unlabeled target data. Although this srategy may boost the adaptation performance, we note that it enforces each unlabeled target data to be assigned the class with the highest predicted probability, which may have two potential risks: (i) some unlabeled target data are aligned to incorrect class centroids since their hard-labels may be incorrect; and (ii) because of issue (i), negative transfer may occur. To alleviate these risks, we propose to adopt the \emph{soft-label} strategy of unlabeled target data instead, which avoids the hard assignment of each unlabeled target data to only one class that may be incorrect. Concretely, the soft-label of the $i$-th projected unlabeled target data $\mathbf{\widetilde{x}}_i^u$, \emph{i.e.}, $\widehat{\mathbf y}_i^u = f(\phi_t(\mathbf{x}_i^u))$, is a $C$-dimensional vector, where the $k$-th element $\widehat{y}_{k,i}^u$ indicates the probability of $\mathbf{\widetilde{x}}_i^u$ belonging to class $k$. We use $\widehat{y}_{k,i}^u$ as the weight of $\mathbf{\widetilde{x}}_i^u$ to calculate the centroid of the $k$-th class in the target domain. Accordingly, a preliminary divergence between class-conditional distributions in the two domains is designed as
\begin{equation}
\small
\label{Q_c'}
\mathcal{Q}_c^{'} \big[ \phi_s(\mathbf{X}_s), \phi_t(\mathbf{X}_t) \big]
= \sum_{k=1}^{C} \left\| \frac{1}{n_s^k}
   \sum_{i=1}^{n_s^k} \mathbf{\widetilde{x}}_{k,i}^s 
   - \frac{\sum_{i=1}^{n_l^k} \widetilde{\mathbf{x}}_{k,i}^l 
         + \sum_{i=1}^{n_u} \widehat{y}_{k,i}^u \widetilde{\mathbf{x}}_i^u}{n_l^k + \sum_{i=1}^{n_u} \widehat{y}_{k,i}^u}
   \right\|^2,   
\end{equation}
where $\mathbf{\widetilde{x}}_{k,i}^{s}$ and $\mathbf{\widetilde{x}}_{k,i}^{l}$ are the $i$-th projected labeled source and target data of class $k$, respectively, and $n^k_s$, $n^k_l$ are the number of labeled source and target data of class $k$, respectively. If $\widehat{y}_{k,i}^u$ adopts the hard-label strategy as in \cite{Long-2013}, then Eq.~(\ref{Q_c'}) reduces to the hard-label approach \cite{Long-2013}. Therefore, the proposed soft-label approach is a generalization of the hard-label approach \cite{Long-2013}.

Besides the \emph{soft-label} strategy introduced in Eq.~(\ref{Q_c'}), we present another \emph{iterative} weighting mechanism to further circumvent the risks mentioned above. We observe that the performance of $f(\cdot)$ gradually improves as the number of iterations increases. Thus, the value of $\widehat{y}_{k,i}^u$ will become more and more precise and reliable as more iterations are performed. As a result, we itroduce an adaptive coefficient to gradually increase the importance  of $\widehat{y}_{k,i}^u$ during adaptation, and based on $\mathcal{Q}_c^{'}[\cdot,\cdot]$ formulate $\mathcal{Q}_c[\cdot,\cdot]$ as
\begin{equation}
\small
\mathcal{Q}_c \big[ \phi_s(\mathbf{X}_s), \phi_t(\mathbf{X}_t) \big]
= \sum_{k=1}^{C} \left\| \frac{1}{n_s^k}
   \sum_{i=1}^{n_s^k} \mathbf{\widetilde{x}}_{k,i}^s
   - \frac{\sum_{i=1}^{n_l^k} \widetilde{\mathbf{x}}_{k,i}^l 
         + \sum_{i=1}^{n_u} \alpha_i^{(r)} \widetilde{\mathbf{x}}_i^u }{n_l^k + \sum_{i=1}^{n_u} \alpha_i^{(r)}} \right\|^2,
\end{equation}
where $\alpha_i^{(r)}$ is defined by 
\begin{equation}
\small
\label{alpha_i}
\alpha_i^{(r)} = \frac{r*\widehat{y}_{k,i}^u}{R},
\end{equation}
which is used to adaptively increase the importance of $\widehat{y}_{k,i}^u$. Here, $R$ is the total number of iterations, and $r$ is the index of current iteration.

\subsection{The Overall Objective of STN}

In summary, to safely transfer knowledge across heterogeneous domains, the ideal model should be able to find the domain-shared classifier, \emph{i.e.}, optimal $f(\cdot)$, and the domain-invariant subspace, \emph{i.e}, optimal $\phi_s(\cdot)$ and $\phi_t(\cdot)$. To this end, we integrate all of the components aforementioned into an end-to-end network, and obtain the overall objective function of STN:
\begin{equation} \label{loss:total}
\small
  \min_{\phi_s, \phi_t, f}
   \mathcal{L}_c \big[ \mathbf{Y}_a, f(\mathbf{X}_a) \big]
   + \beta \mathcal{L}_s \big[ \phi_s(\mathbf{X}_s), \phi_t(\mathbf{X}_t) \big],
\end{equation}
where $\beta$ is a tradeoff parameter to balance the importance between $\mathcal{L}_c [\cdot, \cdot]$ and $\mathcal{L}_s [\cdot, \cdot]$. The proposed STN can jointly learn $f(\cdot)$, $\phi_s(\cdot)$, and $\phi_t(\cdot)$ in an end-to-end network by minimizing Eq.~(\ref{loss:total}). Furthermore, STN adopts both the soft-label and the iterative weighting strategies to match the distributions. These are why STN can be expected to perform quite well as reported in the next section.

\begin{table*}[t]
  \centering
  \setlength{\abovecaptionskip}{2pt}  
  \setlength{\belowrulesep}{0pt}
  \setlength{\aboverulesep}{0pt}
  \setlength\tabcolsep{3.5pt}
  \caption{Classification accuracies (\%) of all the methods on all the image-to-image transfer tasks.}
    \begin{tabular}{c|ccccccccc}
    \toprule
    $\mathcal{D}_s \rightarrow \mathcal{D}_t$ & SVMt  & NNt   & MMDT  & SHFA  & G-JDA & CDLS  & SGW   & TNT   & STN \\
    \midrule
    C$\rightarrow$A  & \multirow{2}[1]{*}{89.13$\pm$0.39} & \multirow{2}[1]{*}{89.6$\pm$0.33} & 87.06$\pm$0.47 & 85.49$\pm$0.51 & 92.49$\pm$0.12 & 86.34$\pm$0.74 & 89.03$\pm$0.37 & 92.35$\pm$0.17 & \textbf{93.03$\pm$0.16} \\
    W$\rightarrow$A  &       &       & 87$\pm$0.47 & 88.83$\pm$0.45 & 92.28$\pm$0.15 & 87.51$\pm$0.44 & 89.02$\pm$0.37 & 92.99$\pm$0.14 & \textbf{93.11$\pm$0.16} \\
    \midrule
    A$\rightarrow$C  & \multirow{2}[0]{*}{79.64$\pm$0.46} & \multirow{2}[0]{*}{81.03$\pm$0.5} & 75.62$\pm$0.57 & 71.16$\pm$0.73 & 86.6$\pm$0.17 & 78.73$\pm$0.49 & 79.88$\pm$0.53 & 85.79$\pm$0.42 & \textbf{88.21$\pm$0.16} \\
    W$\rightarrow$C  &       &       & 75.44$\pm$0.59 & 79.66$\pm$0.52 & 84.82$\pm$0.38 & 77.3$\pm$0.71 & 79.85$\pm$0.53 & 86.28$\pm$0.51 & \textbf{87.22$\pm$0.45} \\
    \midrule
    A$\rightarrow$W  & \multirow{2}[0]{*}{89.34$\pm$0.94} & \multirow{2}[0]{*}{91.13$\pm$0.73} & 89.28$\pm$0.77 & 88.11$\pm$1.01 & 94.09$\pm$0.67 & 91.57$\pm$0.81 & 90.26$\pm$0.84 & 91.26$\pm$0.72 & \textbf{96.68$\pm$0.43} \\
    C$\rightarrow$W  &       &       & 89.11$\pm$0.76 & 89.47$\pm$0.9 & 92.64$\pm$0.54 & 88.6$\pm$0.8 & 90.26$\pm$0.84 & 92.98$\pm$0.75 & \textbf{96.38$\pm$0.38} \\
    \midrule
    A$\rightarrow$D  & \multirow{3}[0]{*}{92.6$\pm$0.71} & \multirow{3}[0]{*}{92.99$\pm$0.63} & 91.65$\pm$0.83 & 95.16$\pm$0.36 & 90.67$\pm$0.65 & 94.45$\pm$0.59 & 93.43$\pm$0.67 & 92.04$\pm$0.76 & \textbf{96.42$\pm$0.43} \\
    C$\rightarrow$D  &       &       & 91.46$\pm$0.85 & 94.25$\pm$0.5 & 88.62$\pm$0.76 & 90.43$\pm$0.79 & 93.43$\pm$0.67 & 92.67$\pm$0.8 & \textbf{96.06$\pm$0.5} \\
    W$\rightarrow$D  &       &       & 91.77$\pm$0.83 & 95.31$\pm$0.63 & 95.87$\pm$0.41 & 92.72$\pm$0.75 & 93.43$\pm$0.67 & 94.09$\pm$0.88 & \textbf{96.38$\pm$0.57} \\
    \midrule
    Avg.  & 87.68$\pm$0.63 & 88.69$\pm$0.55 & 86.49$\pm$0.68 & 87.49$\pm$0.62 & 90.9$\pm$0.43 & 87.52$\pm$0.68 & 88.73$\pm$0.61 & 91.17$\pm$0.44 & \textbf{93.72$\pm$0.36} \\
    \bottomrule
    \end{tabular}%
    \vspace{-3mm}
  \label{tab:image2image}%
\end{table*}%
\begin{table*}[t]
  \centering
  \setlength{\abovecaptionskip}{2pt}  
  \setlength{\belowrulesep}{0pt}
  \setlength{\aboverulesep}{0pt}
  \setlength\tabcolsep{3.5pt}
  \caption{Classification accuracies (\%) of all the methods on the text-to-image transfer task.}
    \begin{tabular}{c|ccccccccc}
    \toprule
    {$\mathcal{D}_s \rightarrow \mathcal{D}_t$} & SVMt  & NNt   & MMDT  & SHFA  & G-JDA & CDLS  & SGW   & TNT   & STN \\
    \midrule
    text$\rightarrow$image & 66.85$\pm$0.96 & 67.68$\pm$0.8 & 53.21$\pm$0.69 & 64.06$\pm$0.61 & 75.76$\pm$0.65 & 70.96$\pm$0.83 & 68.01$\pm$0.8 & 77.71$\pm$0.57 & \textbf{78.46$\pm$0.58} \\
    \bottomrule
    \end{tabular}%
  \label{tab:text2image}%
  \vspace{-3mm}
\end{table*}%

\subsection{Comparison with Existing Studies}

We now compare the proposed STN with some existing studies. To our knowledge, the methods related to STN include MMDT \cite{Hoffman2013,Hoffman-2014}, HFA \cite{Duan-2012}, SHFA \cite{Li-2014}, G-JDA \cite{Hsieh-2016}, CDLS \cite{Tsai-2016}, SGW \cite{Yan-2018}, SCP-ECOC \cite{Xiao-2015a}, JDA \cite{Long-2013}, JGSA \cite{Zhang-2017}, ARTL \cite{Long-2014}, MEDA \cite{Wang-2018}, WDAN \cite{Yan-2017a}, and GAKT \cite{Ding-2018a}. Among these methods, the first seven ones belong to heterogeneous DA techniques, while the last six methods are homogeneous DA ones. However, our work substantially distinguishes from them in the following aspects:
\begin{itemize}
\setlength{\itemsep}{1pt}
\setlength{\parsep}{1pt}
\setlength{\parskip}{1pt}

\item \textbf{Comparison with heterogeneous DA studies}. \textbf{(1)} MMDT, HFA, and SHFA only consider the classifier adaptation. \textbf{(2)} G-JDA, CDLS, and SGW iteratively perform the classifier adaptation and distribution matching. Moreover, G-JDA and CDLS adopt the hard-label strategy to align the conditional distributions, and SGW does not take it into account. \textbf{(3)} SCP-ECOC only matches the marginal distributions.

\item \textbf{Comparison with homogeneous DA studies}. \textbf{(1)} JDA and JGSA take the classifier adaptation and distribution matching as two independent tasks. \textbf{(2)} JDA, JGSA, ARTL, and MEDA utilize the hard-label strategy to reduce the conditional distribution divergence. \textbf{(3)} WDAN estimates the weights of the source data based on the hard-label strategy. \textbf{(4)} GAKT adopts the graph Laplacian regularization to perform the soft-label strategy and neglects the iterative weighting mechanism.

\end{itemize}

\section{Experiments} \label{Experiments}

In this section, we empirically evaluate the proposed STN on the transfer tasks of image-to-image, text-to-image, and text-to-text.

\subsection{Datasets and Settings}

\begin{figure}[t]
  \setlength{\abovecaptionskip}{2pt}
  \centering
  \includegraphics[width=3.2in]{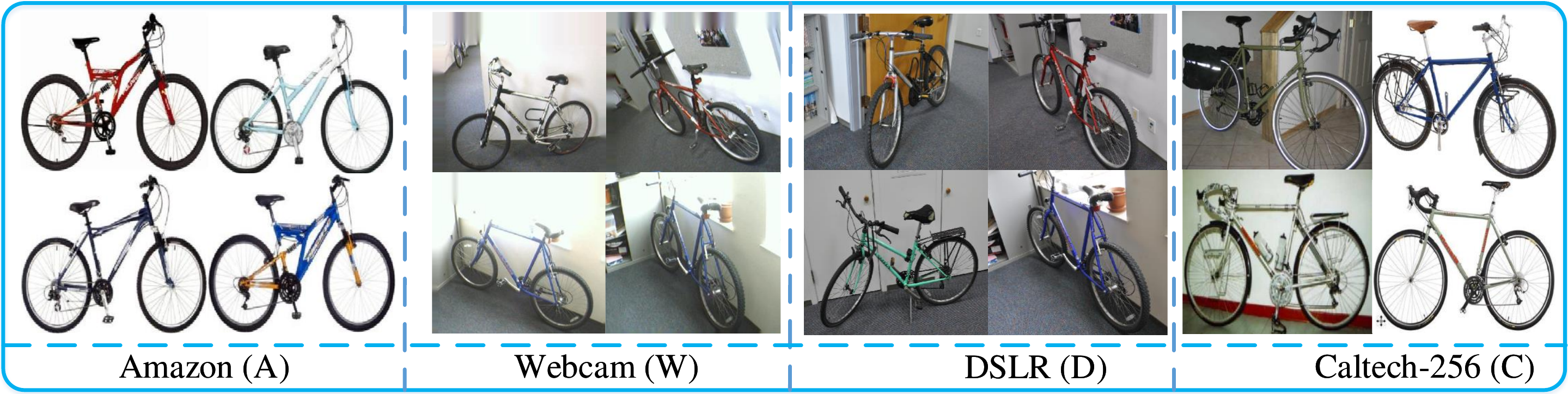}
  \caption{Sample images of the \emph{bike} category from the Office+Caltech-256 dataset.}
  \label{fig:sampleimages}
  \vspace{-4.5mm}
  \end{figure}

The \textbf{image-to-image} transfer task is performed on the \textbf{Office+Cal-tech-256} dataset \cite{Saenko-2010,Griffin-2007}. The former comprises 4,652 images with 31 categories collected from three distinct domains: Amazon (A), Webcam (W), and DSLR (D), and the latter includes 30,607 images of 256 objects from Caltech-256 (C). Following the settings in \cite{Hsieh-2016,Tsai-2016,Chen-2016}, we choose 10 overlapping categories of these two datasets to construct the Office+Caltech-256 dataset. Furthermore, we adopt two different feature representations for this type of task: 800-dimension $SURF$ \cite{Bay-2006} and 4096-dimension $DeCAF_6$ \cite{Donahue-2014}. Some sample images of the category of \emph{bike} are shown in \cref{fig:sampleimages}.

The \textbf{text-to-image transfer} task is conducted on the \textbf{NUS-WIDE+ImageNet} dataset \cite{Chua-2009,Deng-2009}. The former consists of 269,648 images and the associated tags from Flickr, while the latter includes 5,247 synsets and 3.2 million images. We use the tag information of NUS-WIDE (N) and the image data of ImageNet (I) as the domains for text and image, respectively. Following \cite{Chen-2016}, we select 8 shared categories of the two datasets to generate the NUS-WIDE+ImageNet dataset. Moreover, we follow (Chen et al. 2016) to extract the 4-th hidden layer from a 5-layer neural network as the 64-dimensional feature representation for text data, and to extract $DeCAF_6$ feature as the 4096-dimensional feature representation for image data.

The \textbf{text-to-text transfer} task is executed on the \textbf{Multilingual Reuters Collection} dataset \cite{Amini-2009}. This dataset contains about 11,000 articles from 6 categories written in five different languages: English (E), French (F), German (G), Italian (I), and Spanish (S). We follow \cite{Duan-2012,Li-2014,Hsieh-2016,Tsai-2016} to represent all the articles by BOW with TF-IDF features, and then to reduce the dimensions of features by performing PCA with $60\%$ energy preserved. The final dimensions \emph{w.r.t.} E, F, G, I, and S are 1,131, 1,230, 1,417, 1,041, and 807, respectively.

We implement the proposed STN based on the TensorFlow framework \cite{Abadi-2016}. For the sake of fair comparison, we fix the parameter settings of STN for all the tasks. Both $\phi_s(\cdot)$ and $\phi_t(\cdot)$ are two-layer neural networks, which adopt Leaky ReLU \cite{Maas-2013} as the activation function. We use the Adam optimizer \cite{Kingma-2015} with a learning rate of 0.001, and empirically set hyper-parameters $\beta = 0.001, \tau = 0.001$. In addition, the dimension of the common subspace $d$ and the number of iterations $R$ are set to 256 and 300, respectively.

\subsection{Evaluations}

 We evaluate the proposed STN against eight state-of-the-art supervised learning and HDA methods, \emph{i.e.}, SVMt, NNt, MMDT \cite{Hoffman2013,Hoffman-2014}, SHFA \cite{Li-2014}, G-JDA \cite{Hsieh-2016}, CDLS \cite{Tsai-2016}, SGW \cite{Yan-2018}, and TNT \cite{Chen-2016}. Among these methods, SVMt and NNt train a support vector machine and a neural network with only the labeled target data, respectively. MMDT, SHFA, G-JDA, CDLS, and SGW are the shallow HDA methods, while TNT is the deep HDA one. Furthermore, we note that SHFA, G-JDA, CDLS, SGW, and TNT are semi-supervised HDA methods, which learn from both labeled and unlabeled cross-domain data. As stated in Section \ref{RelatedWork}, homogeneous DA methods cannot directly apply to the HDA problem, thus we do not include them in the comparison. Following \cite{Li-2014,Hsieh-2016,Chen-2016,Tsai-2016,Yan-2018}, we use the classification accuracy as the evaluation metric.

\begin{figure*}[t]
\setlength{\abovecaptionskip}{3pt}
\centering
\subfigure[E$\rightarrow$S] {
{\includegraphics[width=0.456\columnwidth]{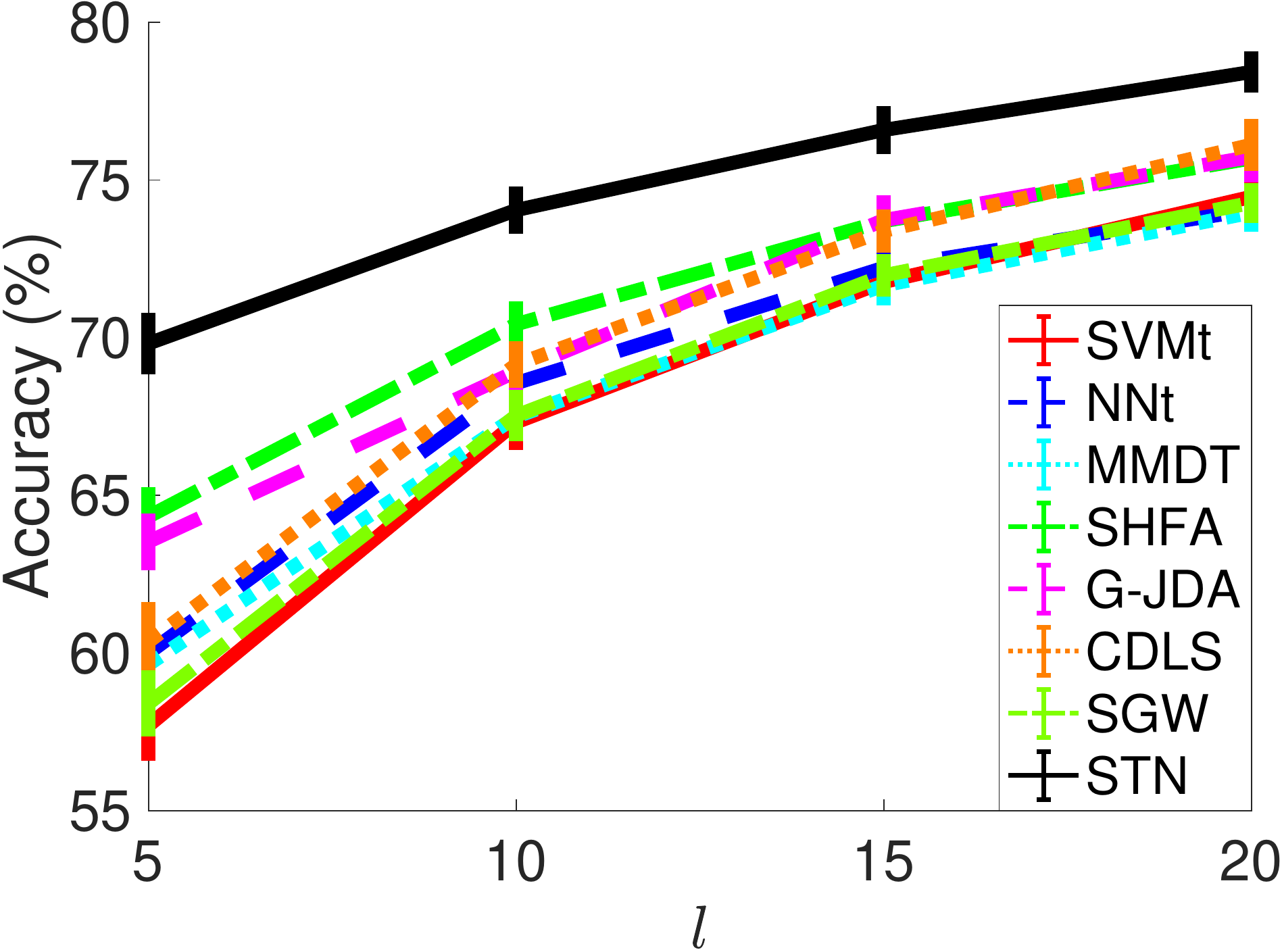}}
}
\subfigure[F$\rightarrow$S] {
{\includegraphics[width=0.456\columnwidth]{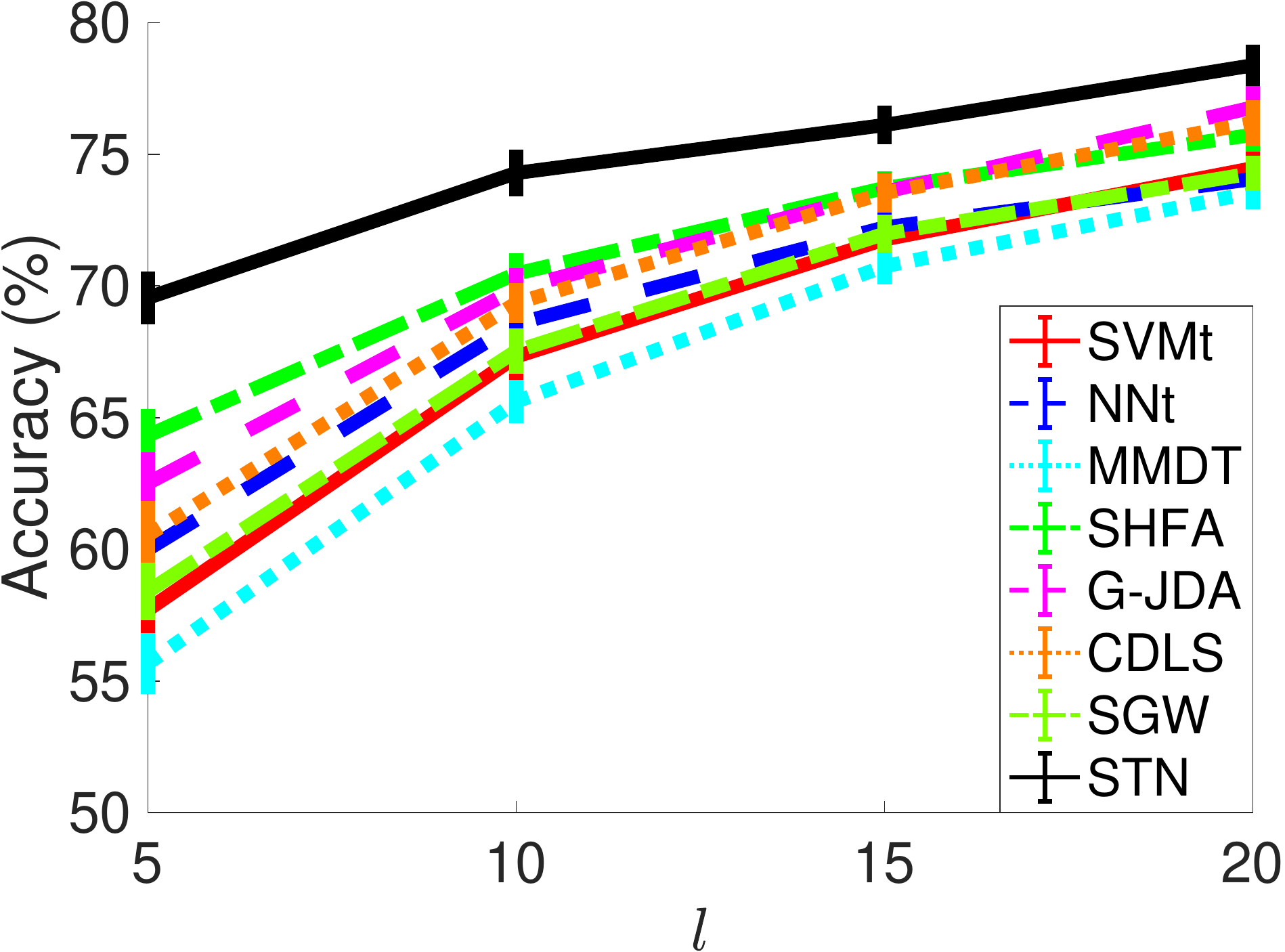}}
}
\subfigure[G$\rightarrow$S] {
{\includegraphics[width=0.456\columnwidth]{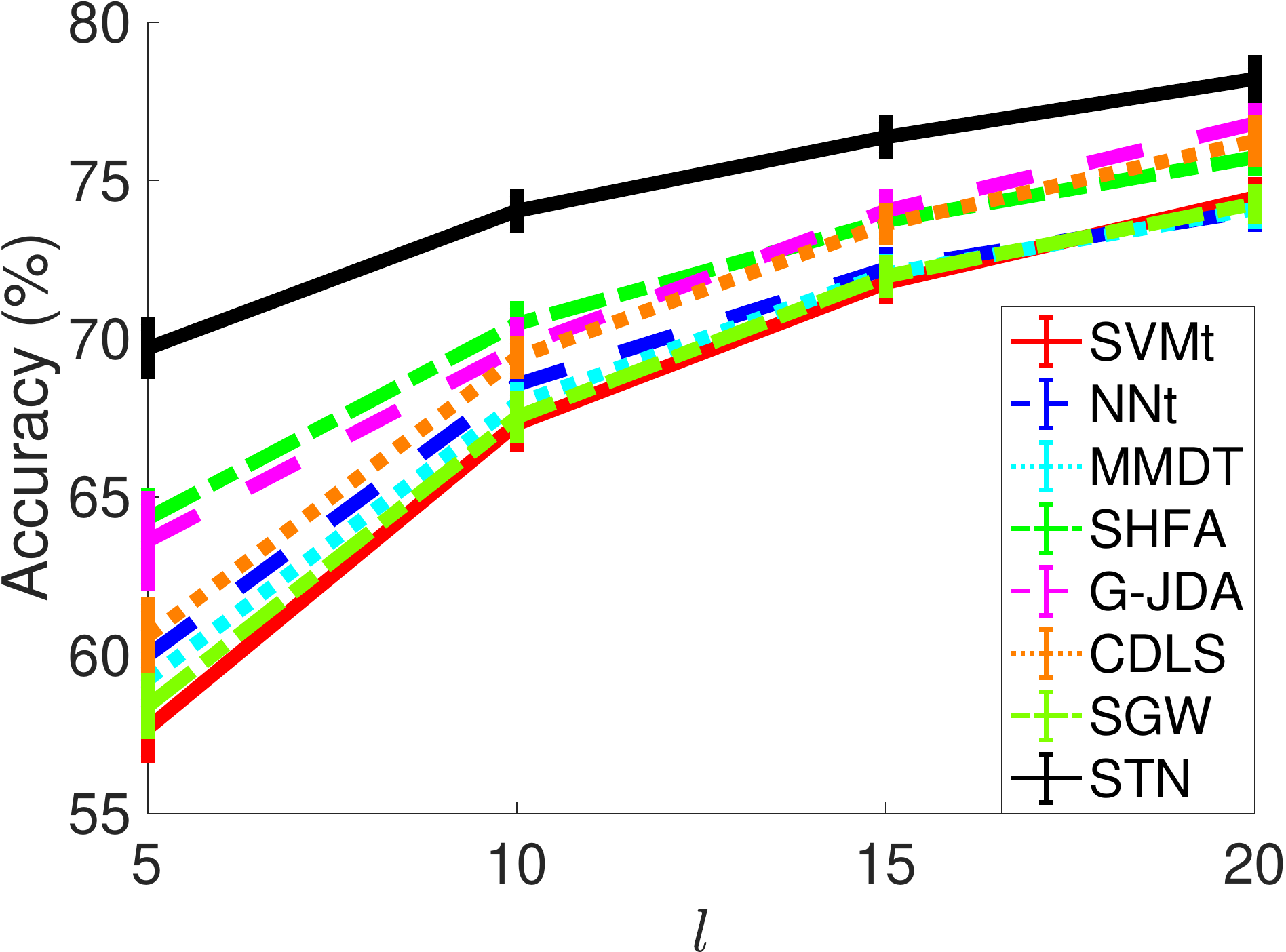}}
}
\subfigure[I$\rightarrow$S] {
{\includegraphics[width=0.456\columnwidth]{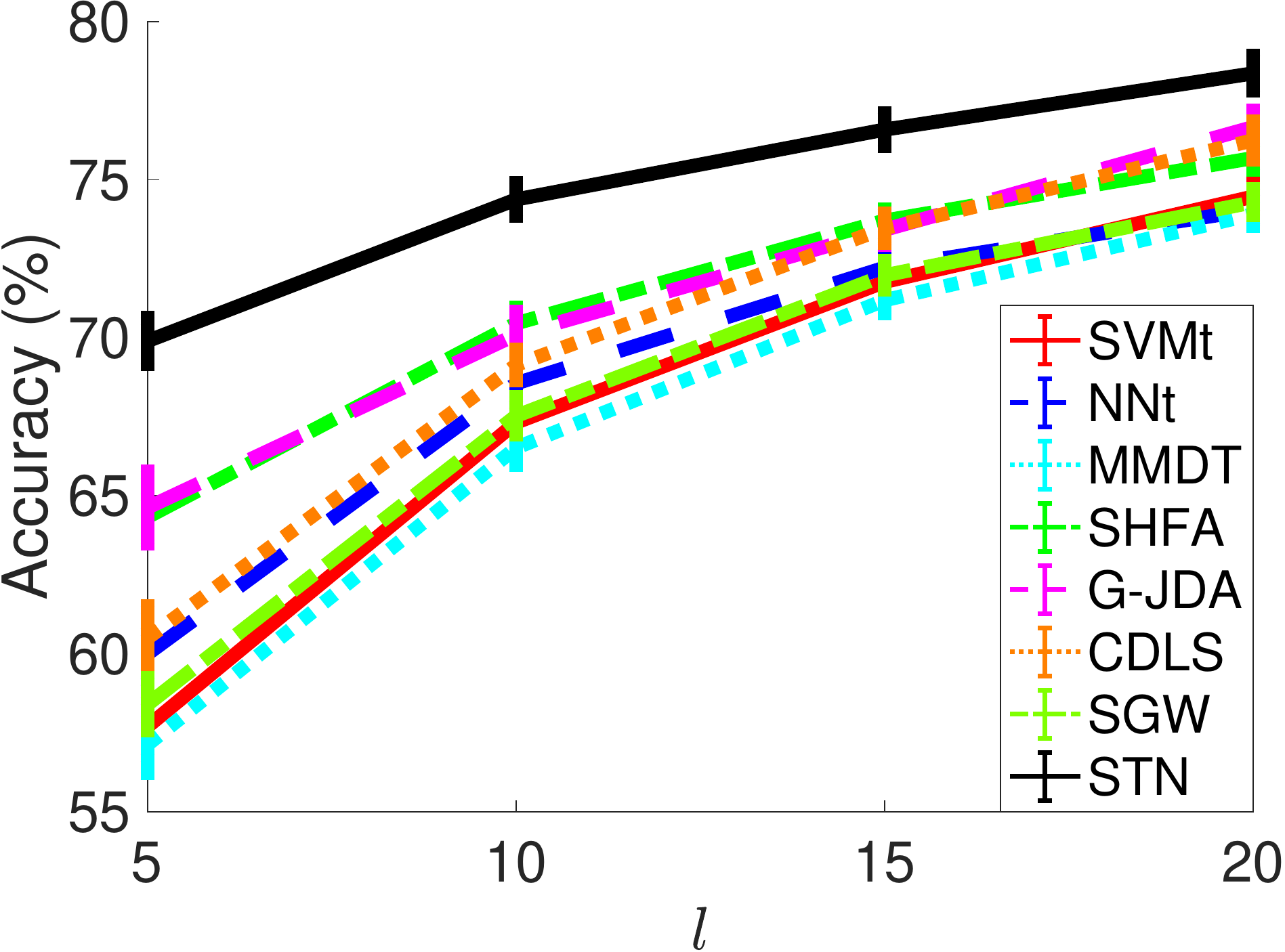}}
}
\caption{Classification accuracies (\%) of all the methods with different numbers of labeled target data per category (\emph{i.e.}, $l = 5,10,15,20$) on all the text-to-text transfer tasks.}
\label{fig:text2text}
\vspace{-5mm}
\end{figure*}

\noindent \textbf{Image-to-image transfer:} We first perform the task of image-to-image transfer on the Office+Caltech-256 dataset. In this kind of task, the source and target data are drawn from not only different feature representations but also distinct domains. As for the source domain, we take images in $SURF$ features and utilize all images as the labeled data. For the target domain, we represent images with $DeCAF_6$ features and randomly sample 3 images per class as the labeled data. The remaining images in the target domain are used as the testbed. Moreover, D is only viewed as the target domain due to the limited amount of images. \cref{tab:image2image} presents the average classification accuracies of 20 random trials. 

From the results, we can make several insightful observations. \textbf{(1)} The proposed STN consistently achieves the highest accuracies on all the tasks. The average classification accuracy of STN is \textbf{93.72\%}, which makes the improvement over the best supervised learning method, \emph{i.e.}, NNt, and the best HDA method, \emph{i.e.}, TNT, by \textbf{5.03\%} and \textbf{2.55\%}, respectively. These results clearly demonstrate the superiority of STN. \textbf{(2)} STN performs significantly better than MMDT and SHFA. The reason is that they only consider the classifier adaptation strategy but neglect the distribution matching strategy which can reduce the distributional divergence across domains. Although G-JDA and CDLS combine both strategies for HDA, their performance is still worse than STN. One reason is because they iteratively perform the strategies of classifier adaptation and distribution matching. The iterative combination is a bit heuristic and may lead to unstable performance. Another important reason is that they use the hard-label strategy of unlabeled target data during the distribution matching. The hard-labels may be not correct, which may result in limited performance improvement and negative transfer. The performance of SGW is worse than STN. An important reason is that SGW does not utilize unlabeled target data to align the conditional distributions between domains. STN yields better performance than TNT with one reason that TNT does not explicitly minimize the distributional discrepancy between domains. \textbf{(3)} MMDT and SHFA do not always outperform the supervised learning method, \emph{i.e.}, SVMt. One possible explanation is that the distributional divergence across domains may be large. Although they both train a domain-shared classifier to align the discriminative directions of domains, they do not minimize the distributional difference very well due to the limited amount of the labeled target data. Thus, it is risky to overfit the target data that result in negative transfer. In addition, SHFA exceeds MMDT with one major reason that the former utilizes the unlabeled target data while the latter does not. \textbf{(4)} G-JDA, CDLS, and SGW are better or comparable than SVMt. They iteratively perform the strategies of classifier adaptation and distribution matching, which confirms that these strategies are both meaningful and useful. In addition, we note that CDLS achieves worse performance than G-JDA. A possible explanation is that the instance weighting scheme used in CDLS is not always beneficial and may hurt the performance. \textbf{(5)} STN and TNT perform better than all the other methods, which verifies deep models are effective for addressing the HDA problem in the image domain.

\noindent \textbf{Text-to-image transfer:} We then conduct the text-to-image transfer task on the NUS-WIDE+ImageNet dataset. It is very hard because there is no co-occurrence text and image data for learning. According to \cite{Chen-2016}, we treat the text (\emph{i.e.}, N) and image (\emph{i.e.}, I) datasets as the source and target domains, respectively. For the source domain, we choose 100 texts per category as the labeled data. As for the target domain, we randomly sample 3 images as the labeld data from each class, and the rest images are considered as the testbed. \cref{tab:text2image} shows the average classification accuracies in 20 random trials. 

From the results, we can make the following important observations. \textbf{(1)} The proposed STN achieves the best performance on this task. The classification accuracy of STN is \textbf{78.46\%}, which outperforms the best supervised learning method, \emph{i.e.}, NNt, and the best HDA method, \emph{i.e.}, TNT, by \textbf{10.78\%} and \textbf{0.75\%}, respectively. These results further corroborate the effectiveness of STN. \textbf{(2)} Both STN and TNT significantly exceed the other methods, which again verifies deep networks is helpful for learning heterogeneous cross-domain data. \textbf{(3)} The performance of MMDT is quite poor. This observation is consistent with \cite{Chen-2016} with one possible reason that the distributional divergence between domains may be very large since the source domain is textual while the target one is visual. \textbf{(4)} We have the same observation as the image-to-image transfer task that SHFA and G-JDA are better than MMDT and CDLS, respectively.

  \begin{figure}[t]
  \setlength{\abovecaptionskip}{2pt}
  \centering
  {\includegraphics[width=0.9\columnwidth]{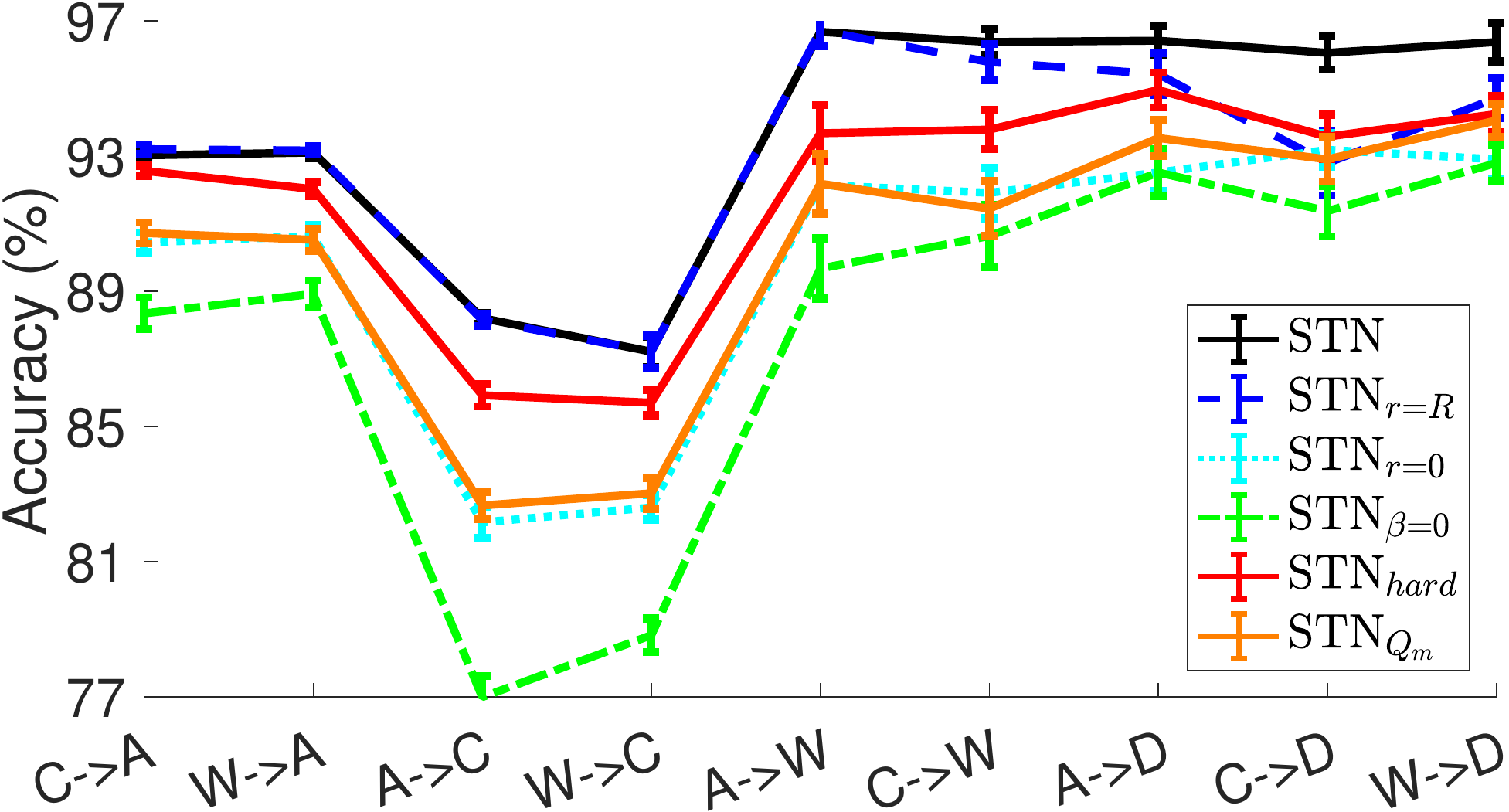}}
  \caption{Performance comparison of STN and its variants on all the image-to-image transfer tasks. Here, STN$_{r=R}$ removes the iterative weighting mechanism, STN$_{r=0}$ ignores the unlabeled target data, STN$_{\beta=0}$ ablates the soft-MMD loss, STN$_{hard}$ adopts the hard-label strategy of unlabeled target data, and STN$_{Q_m}$ neglects the divergence between conditional distributions.}
  \label{fig:variants}
  \vspace{-4.5mm}
  \end{figure}
\begin{figure*}[t]
\setlength{\abovecaptionskip}{2pt}
\centering
\subfigure[C with $SURF$ \label{fig:Tsne_source_C}] {
{\includegraphics[width=0.456\columnwidth]{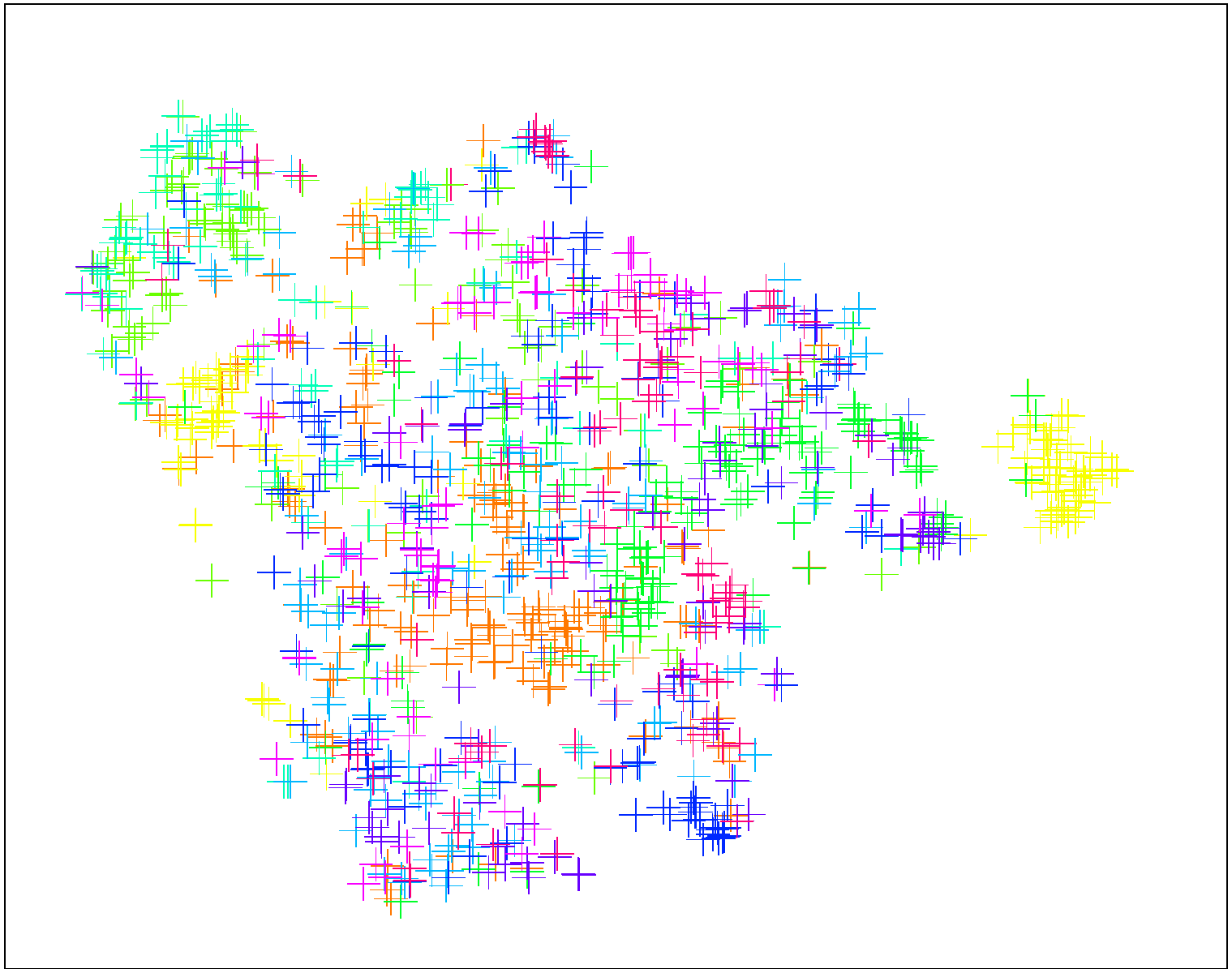}}
}
\subfigure[W with $DeCAF_6$ \label{fig:Tsne_target_W}] {
{\includegraphics[width=0.456\columnwidth]{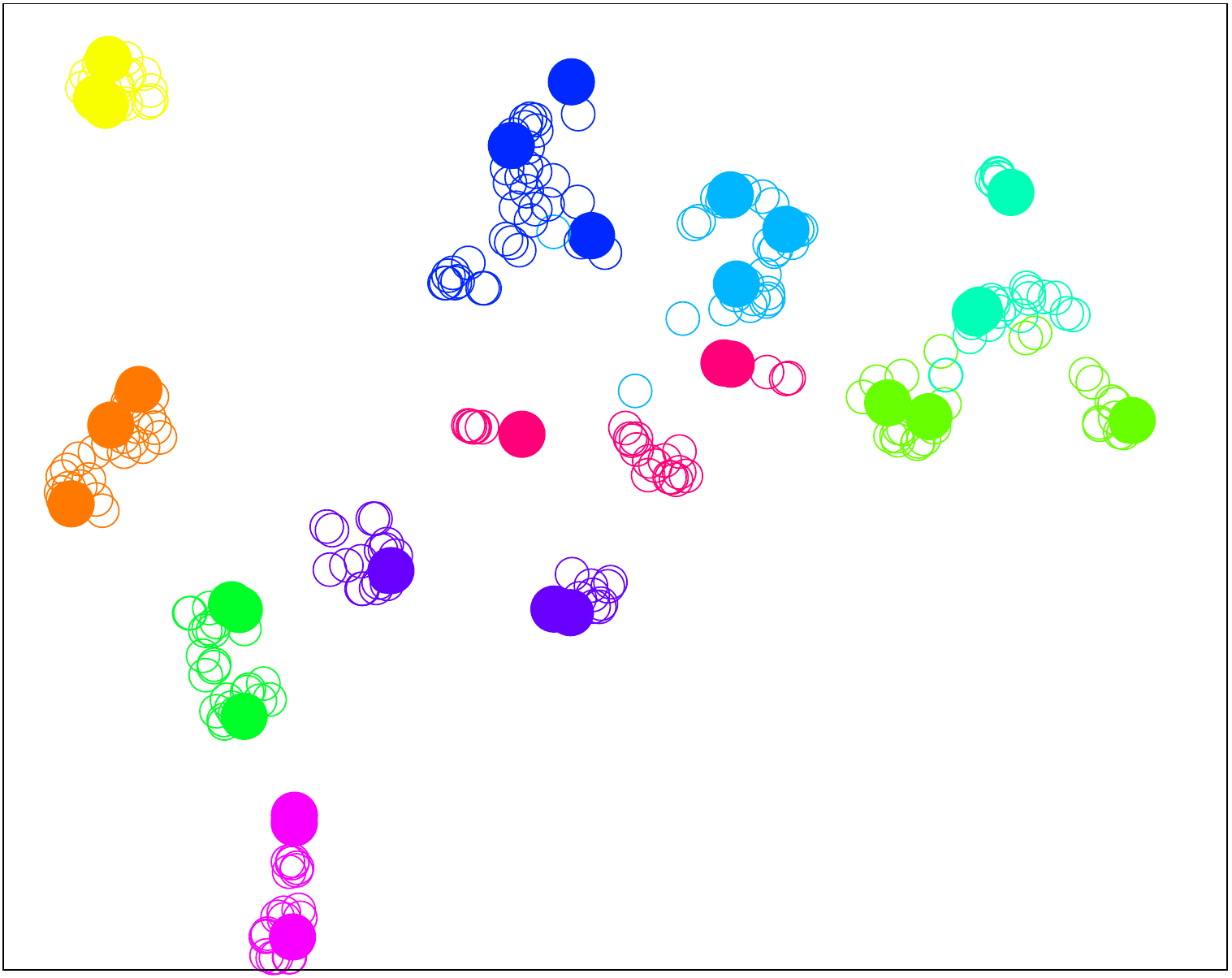}}
}
\subfigure[MMDT \label{fig:Tsne_MMDT}] {
{\includegraphics[width=0.456\columnwidth]{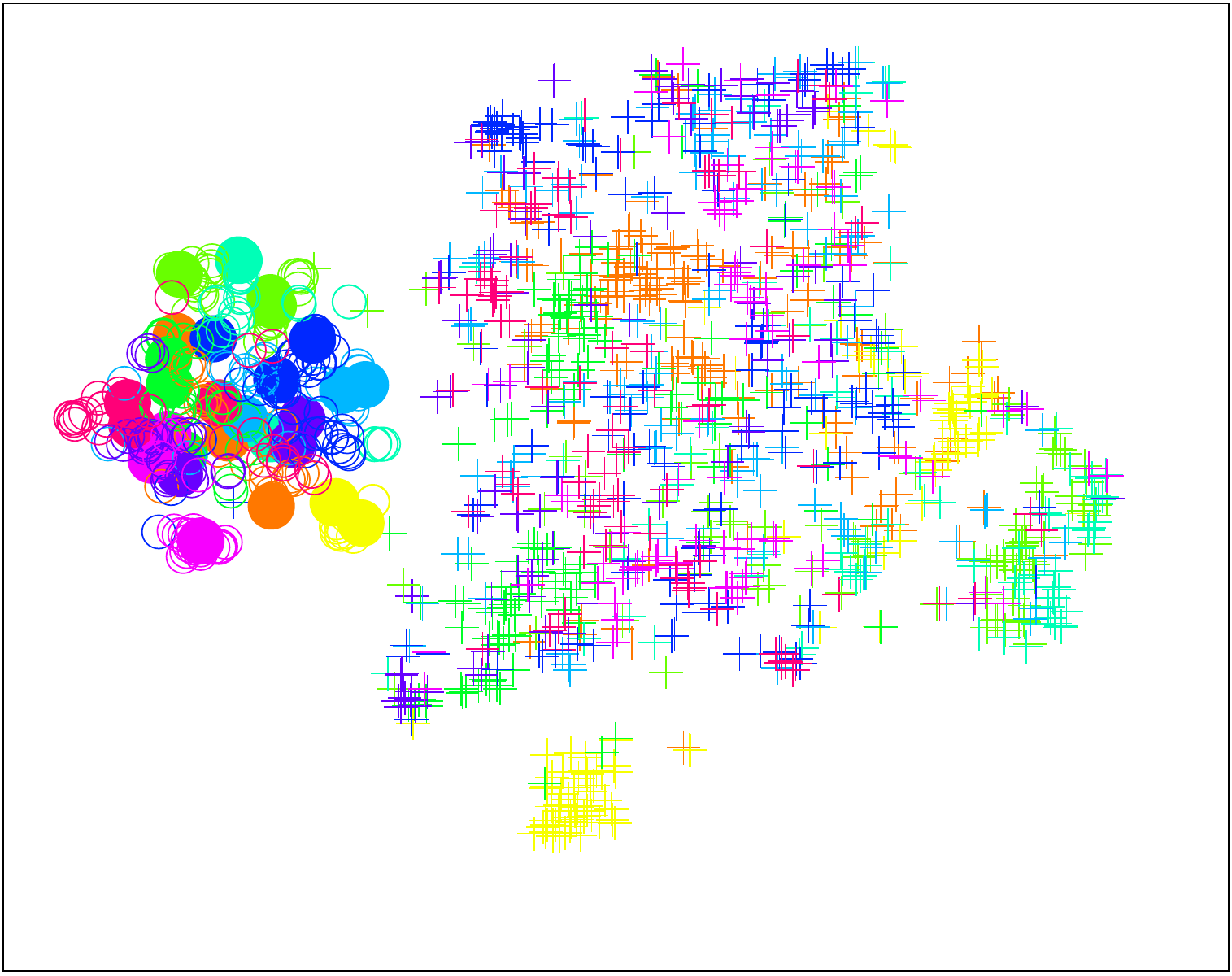}}
}
\subfigure[G-JDA \label{fig:Tsne_GJDA}] {
{\includegraphics[width=0.456\columnwidth]{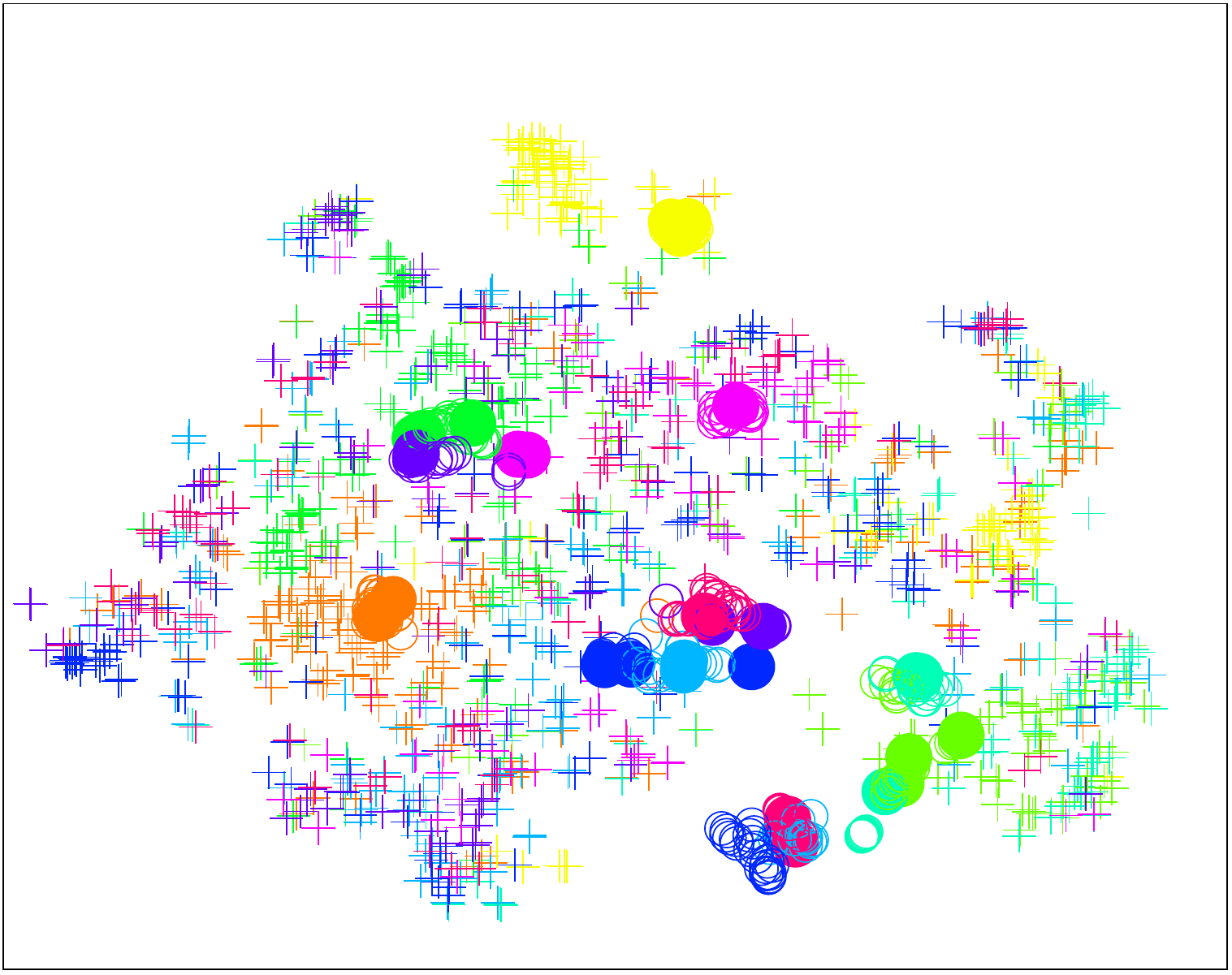}}
}
\subfigure[CDLS \label{fig:Tsne_CDLS}] {
{\includegraphics[width=0.456\columnwidth]{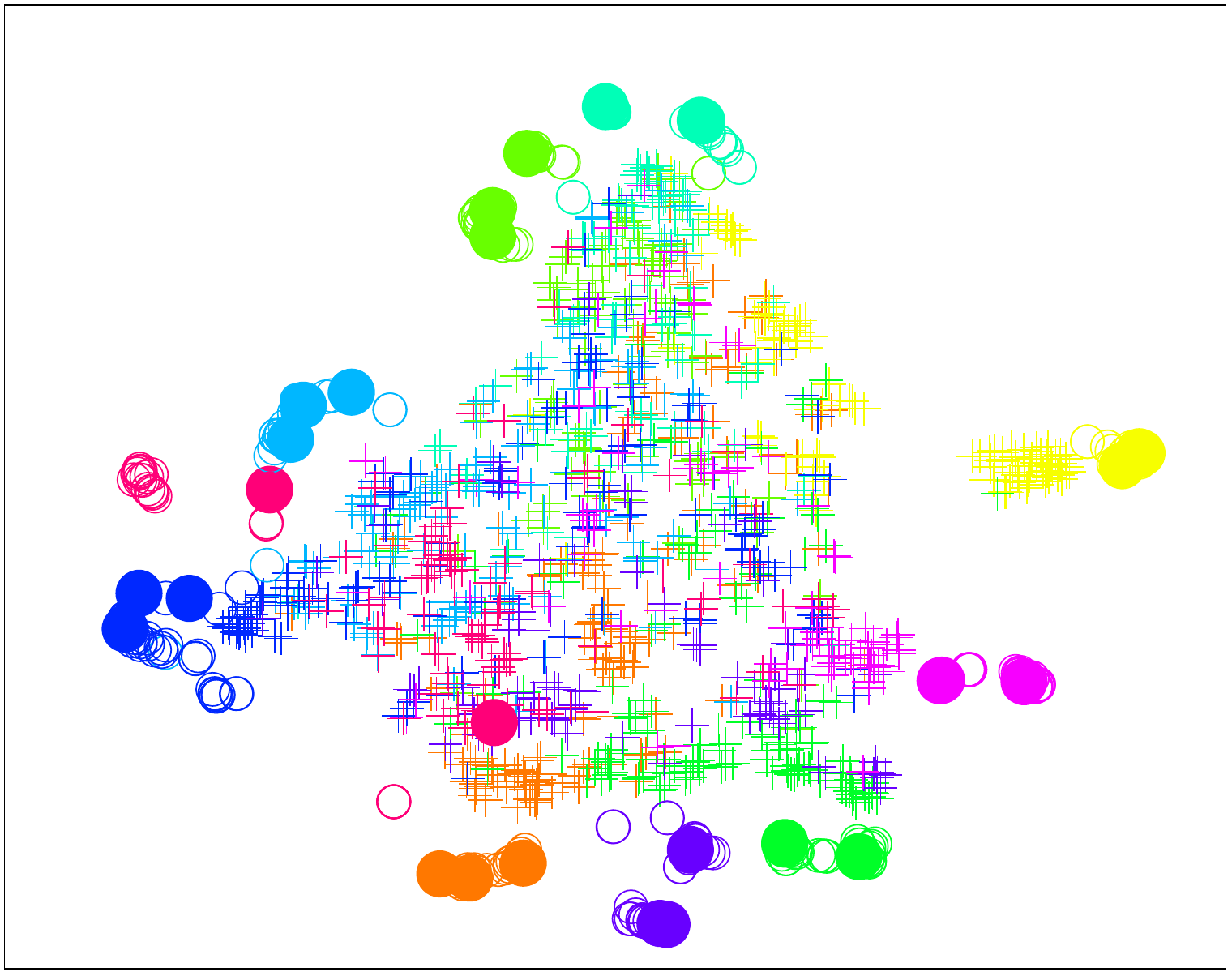}}
}
\subfigure[SGW \label{fig:Tsne_SGW}] {
{\includegraphics[width=0.456\columnwidth]{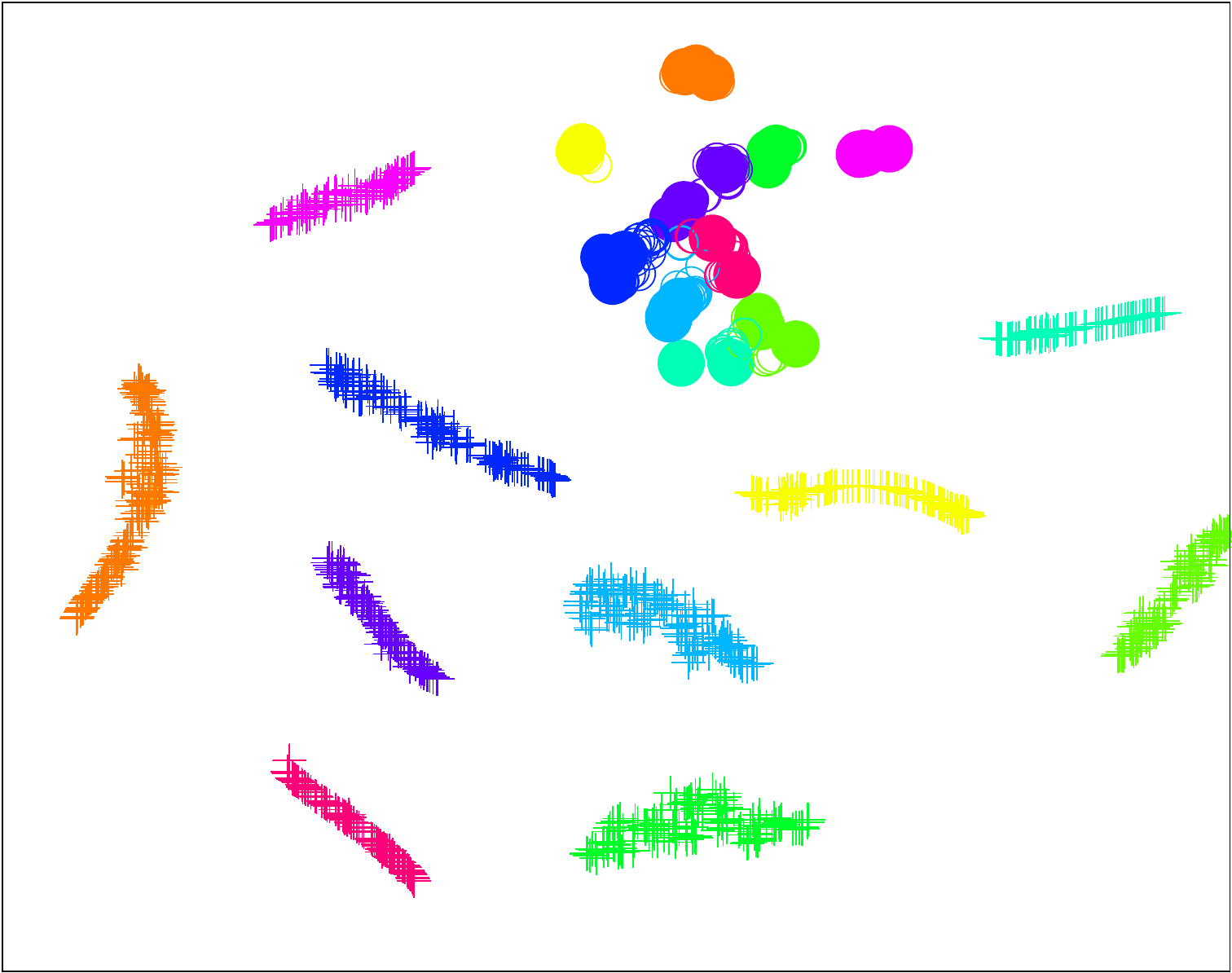}}
}
\subfigure[TNT \label{fig:Tsne_TNT}] {
{\includegraphics[width=0.456\columnwidth]{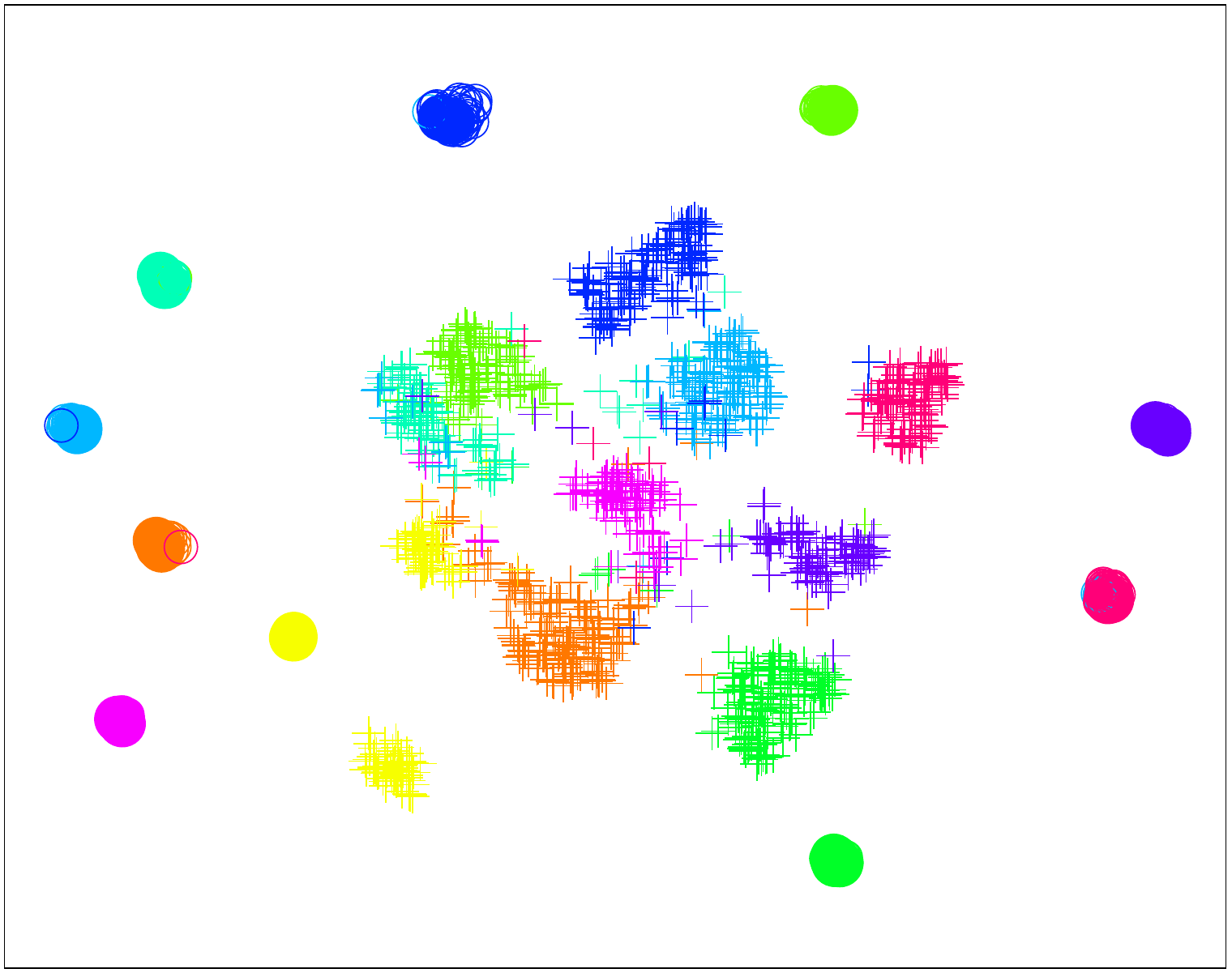}}
}
\subfigure[STN \label{fig:Tsne_STN}] {
{\includegraphics[width=0.456\columnwidth]{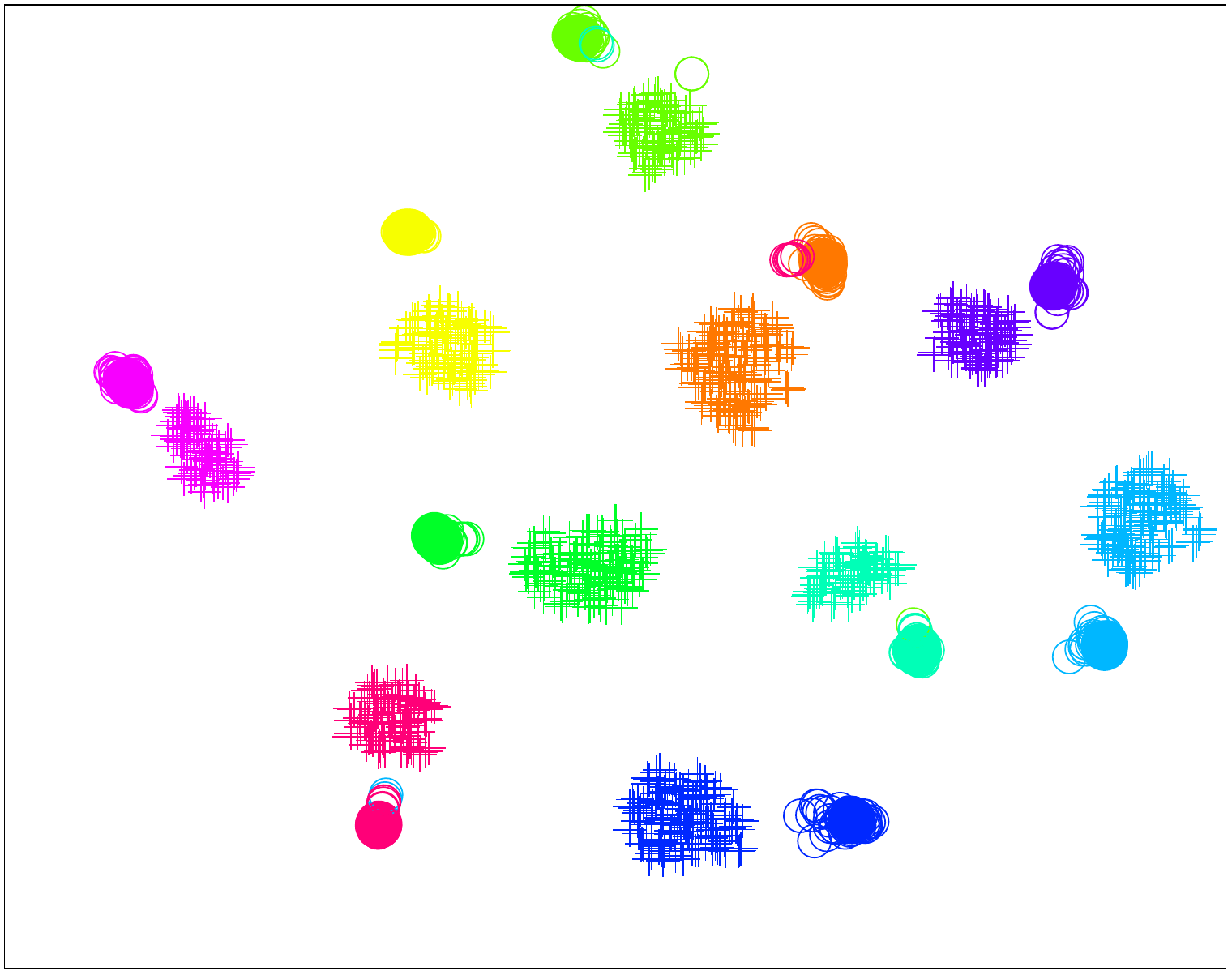}}
}
\includegraphics[width=1.\columnwidth]{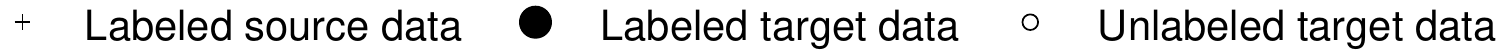}
\caption{The t-SNE visualization on the C$\rightarrow$W task. Here, (a) and (b) are the original feature representations of $SURF$ and $DeCAF_6$, respectively. (c), (d), (e), (f), (g), and (h) are the learned feature representations of MMDT, G-JDA, CDLS, SGW, TNT, and STN, respectively.}
\label{fig:visualization}
\vspace{-5mm}
\end{figure*}

\noindent \textbf{Text-to-text transfer:} We now execute the text-to-text transfer task on the Multilingual Reuters Collection dataset. In this type of task, we regard articles in distinct languages as data in different domains. According to the settings in \cite{Duan-2012,Li-2014,Hsieh-2016,Tsai-2016}, we consider E, F, G and I as the source domains, and S as the target one. For the source domain, we randomly pick up 100 articles per category as the labeled data. As for the target domain, we randomly select $l$ (\emph{i.e.}, $l=5,10,15,20$) and 500 articles per class as the labeled and unlabeled data, respectively. We explore how the number of labeled target data per category $l$ affects the performance and plot the average classification accuracies of 20 random trials in \cref{fig:text2text}. We do not report the results of TNT as they are much worse than the other methods (we note that the original paper of TNT also does not report the results on this type of task, please refer to \cite{Chen-2016} for details). 

From the results, we have several interesting observations. \textbf{(1)} The performance of all the methods are boosted with the increase of the number of labeled target data. This observation is intuitive and reasonable. \textbf{(2)} The proposed STN substantially outperforms all the baseline methods on all the tasks. In particular, the average classification accuracy of STN with 5 labeled target data on all the tasks is \textbf{69.75\%}, which outperforms the best supervised learning method, \emph{i.e.}, NNt, and the best HDA method, \emph{i.e.}, SHFA, by \textbf{9.75\%} and \textbf{5.42\%}, respectively. These results verify the effectiveness of STN again. \textbf{(3)} All the HDA methods can yield comparable or better performance than the supervised learning method, \emph{i.e.}, SVMt, which implies that these methods can produce positive transfer on all the text-to-text transfer tasks. \textbf{(4)} SHFA and G-JDA outperform MMDT and CDLS, respectively. We have the same observation as that on the image-to-image and text-to-image transfer tasks.

\begin{figure*}[t]
\setlength{\abovecaptionskip}{2pt}
\centering
\subfigure[$\beta$ \label{fig:beta}] {
{\includegraphics[width=0.456\columnwidth]{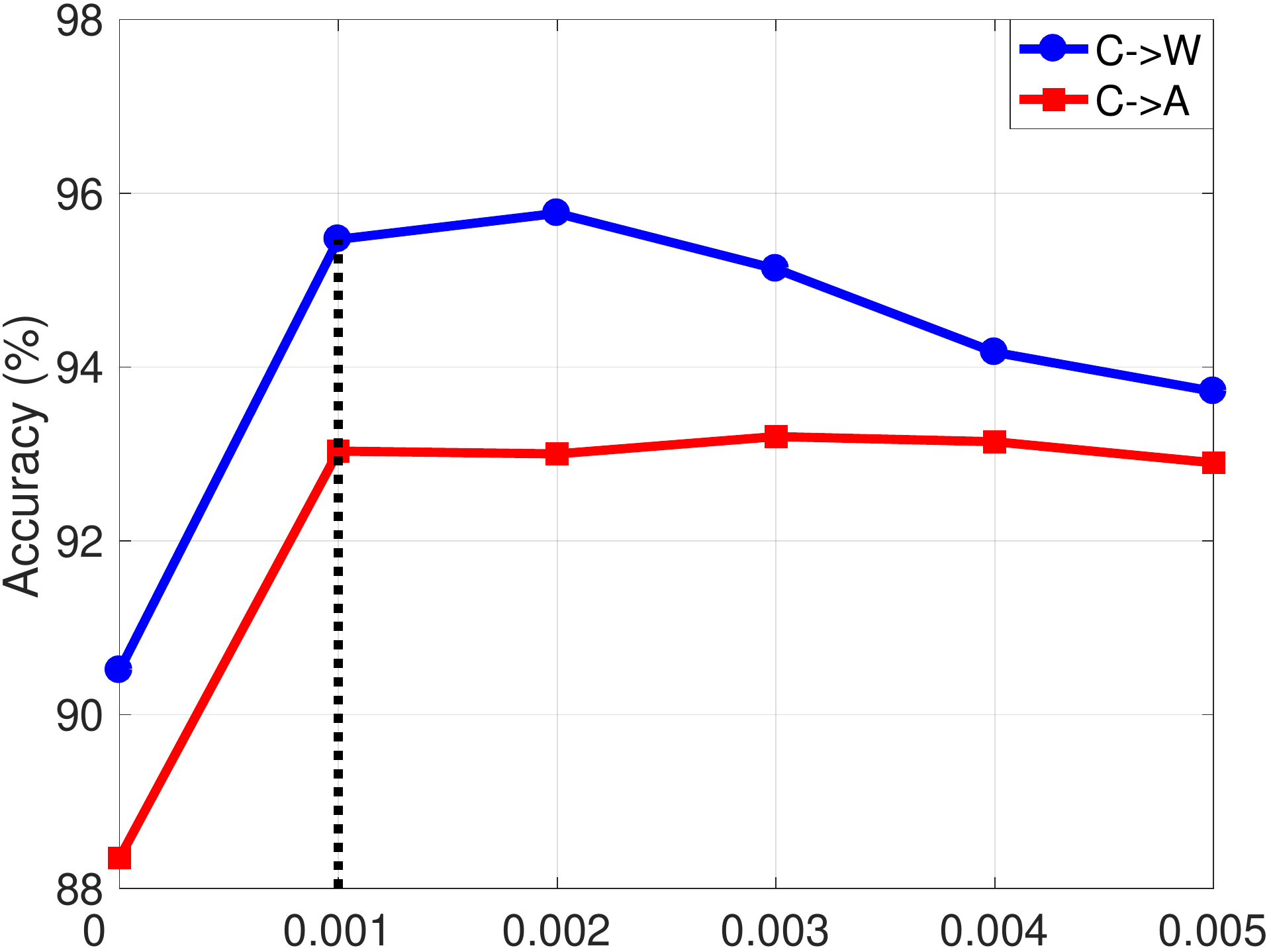}}
}
\subfigure[$\tau$ \label{fig:tau}] {
{\includegraphics[width=0.456\columnwidth]{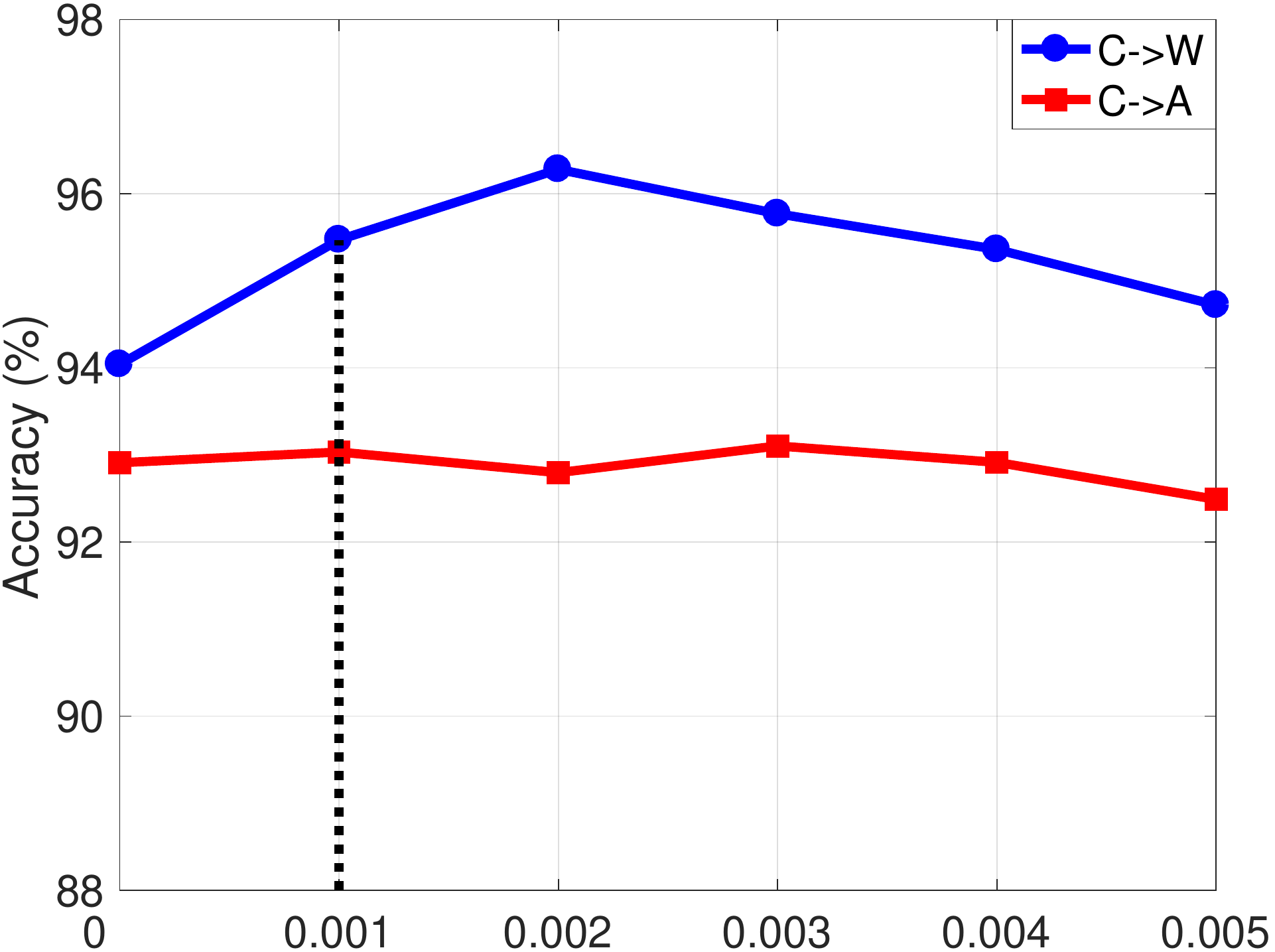}}
}
\subfigure[$d$ \label{fig:d}] {
{\includegraphics[width=0.456\columnwidth]{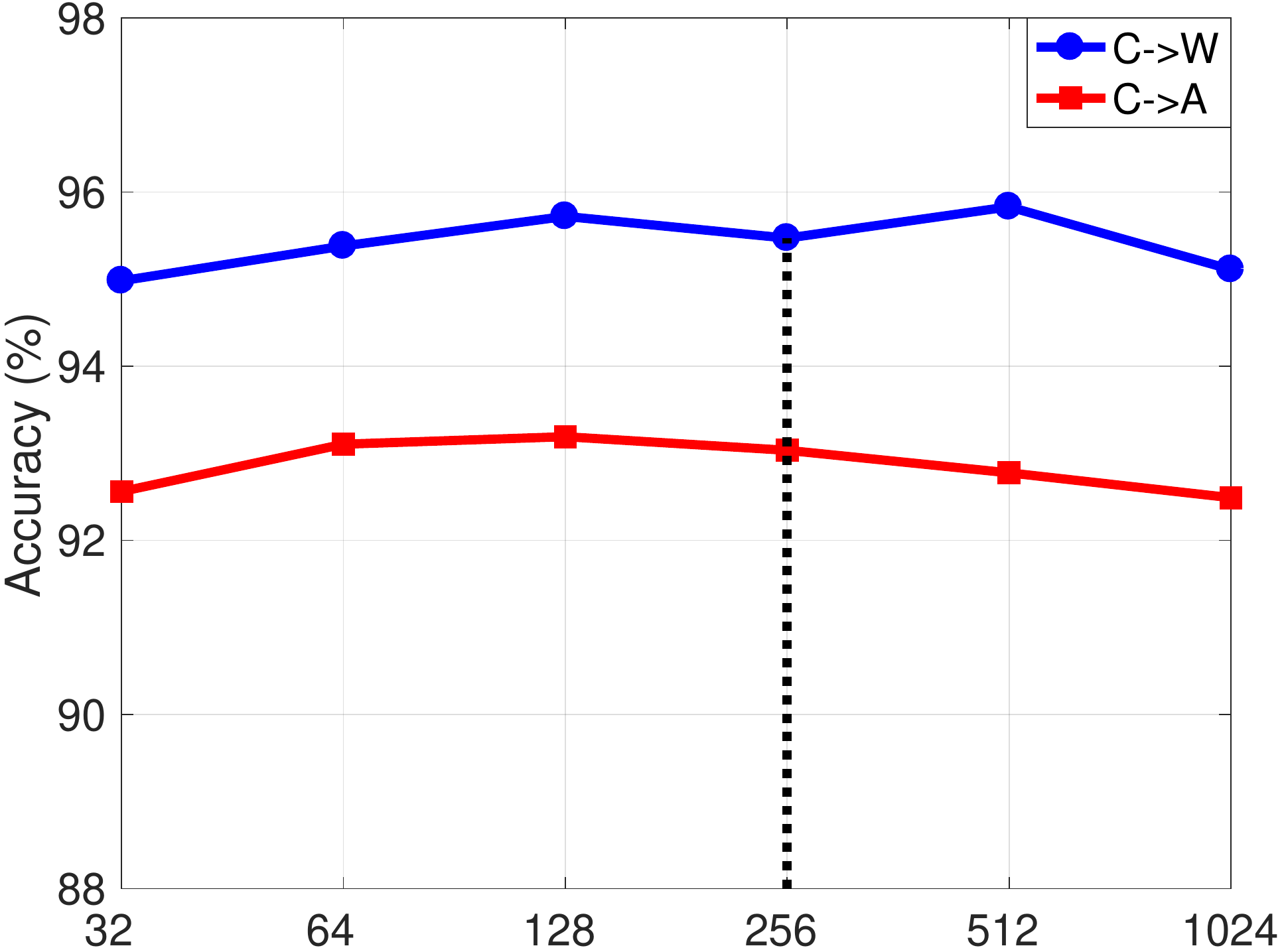}}
}
\subfigure[$R$ \label{fig:convergence}] {
{\includegraphics[width=0.456\columnwidth]{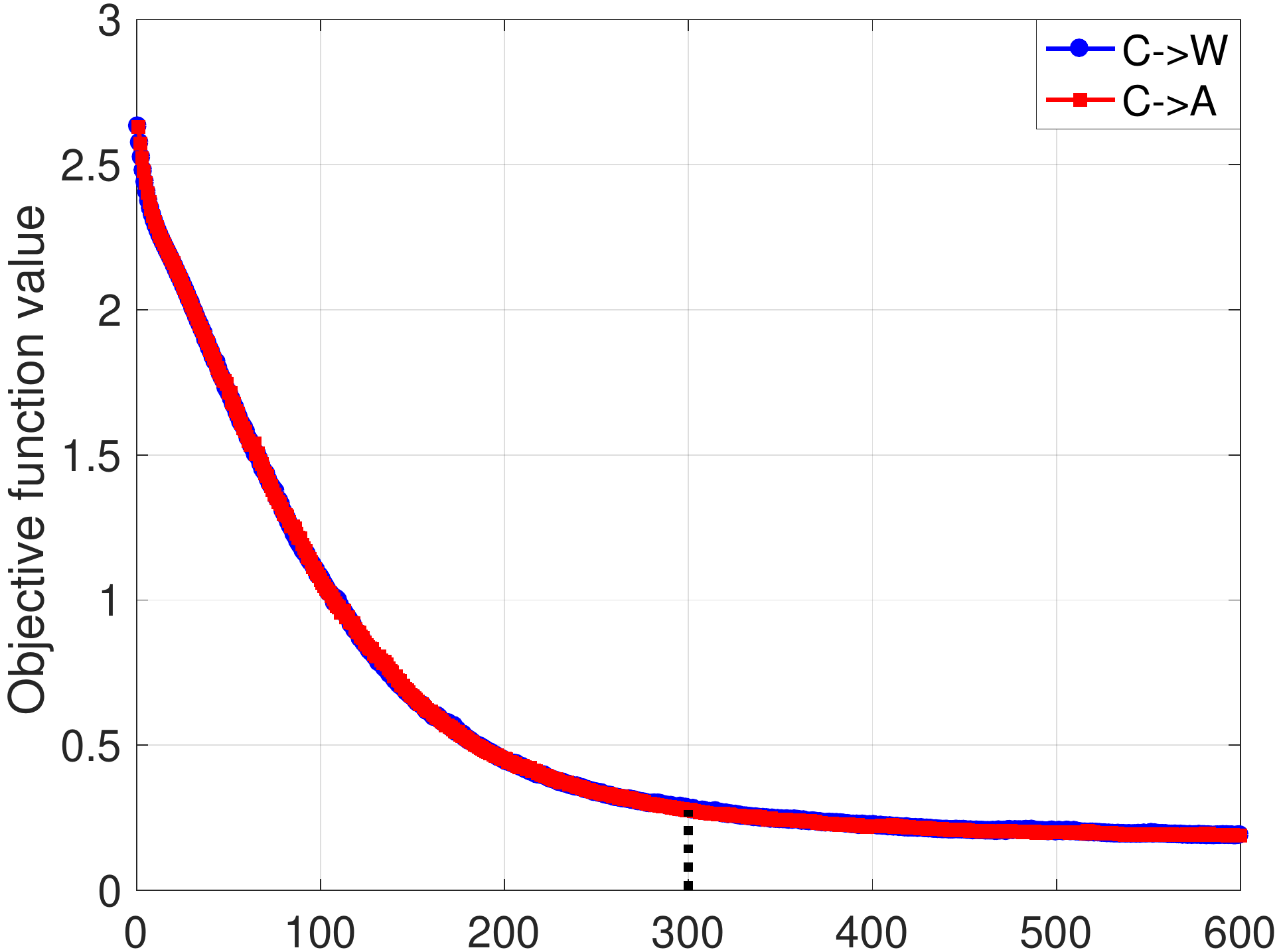}}
}
\caption{Parameter sensitivity and convergence analysis on the transfer tasks of C$\rightarrow$W and C$\rightarrow$A.}
\label{fig:analysis}
\vspace{-5mm}
\end{figure*}

\subsection{Analysis}

\textbf{Variants Evaluation:} To go deeper with the efficacy of the iterative weighting mechanism, the distribution matching strategy, and the soft-label strategy, we evaluate several variants of STN: \textbf{(1)} STN$_{r=R}$, which removes the iterative weighting mechanism by setting $r = R$ in Eq.~(\ref{alpha_i}); \textbf{(2)} STN$_{r=0}$, which ignores the unlabeled target data by setting $r = 0$ in Eq.~(\ref{alpha_i}); \textbf{(3)} STN$_{\beta=0}$, which ablates the soft-MMD loss by setting $\beta=0$ in Eq.~(\ref{loss:total}); \textbf{(4)} STN$_{hard}$, which adopts the hard-label strategy of unlabeled target data and iteratively performs the hard-label assignment and objective optimizing in an end-to-end network; and \textbf{(5)} STN$_{Q_m}$, which neglects the divergence on the conditional distributions by ablating $Q_c$ in Eq.~(\ref{loss:SMMD}). We use the same experimental setting as above and plot the average classification accuracies on all the image-to-image transfer tasks in \cref{fig:variants}. The results reveal several insightful observations. \textbf{(1)} As expected, STN delivers the best performance on all the tasks. \textbf{(2)} STN$_{r=R}$ is worse than STN, which suggests that the iterative weighting mechanism is useful to further improve the performance. STN$_{hard}$ is worse than STN and STN$_{r=R}$, which implies that the soft-label strategy is more suitable to match the conditional distributions between domains than the hard-label. STN$_{r=0}$ is worse than STN, STN$_{r=R}$, and STN$_{hard}$, which indicates that using the unlabeled target data can further improve the performance without resulting in negative transfer on these tasks. STN$_{Q_m}$ is worse than STN, STN$_{r=R}$, and STN$_{hard}$, which indicates that aligning the conditional distributions is useful. STN$_{Q_m}$ is similar to STN$_{r=0}$, which implies that using the unlabeled target data is important for reducing the divergence on conditional distributions. STN$_{\beta=0}$ yields the worst performance, which suggests that the distribution matching strategy is necessary and helpful for transferring knowledge across heterogeneous domains.

\noindent \textbf{Feature Visualization:} We use the t-SNE technique \cite{Maaten-2014} to visualize the learned features of all the methods on the task of C$\rightarrow$W except SHFA because it does not explicitly learn the feature projection matrices (please refer to \cite{Li-2014} for details). We plot the visualization results in \cref{fig:visualization}. The results offer several interesting observations. \textbf{(1)} Comparing \cref{fig:Tsne_source_C} with \cref{fig:Tsne_target_W}, we can see that the discriminative ability of $DeCAF_6$ is better than $SURF$, which is reasonable as $DeCAF_6$ is the deep feature. \textbf{(2)} For MMDT, G-JDA, CDLS, SGW, and TNT, we can find that they do not align the distributions between domains very well, which explains their poor performance on this task. \textbf{(3)} The proposed STN matches the distributions between domains very well, which indicates that STN is powerful for transferring knowledge across heterogeneous domains.

\noindent \textbf{Parameter Sensitivity and Convergence:} We conduct experiments to analyze the parameter sensitivity and the convergence of STN on the transfer tasks of C$\rightarrow$W and C$\rightarrow$A. Figures~\ref{fig:beta}-\ref{fig:d} show the accuracy \emph{w.r.t.} $\beta$, $\tau$, and $d$, respectively. We can observe that the default values ($\beta=0.001, \tau=0.001, d=256$) can achieve high accuracies. It is worth noting that STN yields the state-of-the-art performance on all the tasks by taking these default parameter values, which indicates STN is quite stable and effective. In addition, we plot the objective function value \emph{w.r.t.} the number of iterations in \cref{fig:convergence}. We can see that the value of objective function first decreases and then tends to become steady as more iterations are executed, indicating the convergence of STN.

\section{Conclusion} \label{Conclusion}

This paper proposes a STN to address the HDA problem, which jointly learns a domain-shared classifier and a domain-invariant subspace in an end-to-end way. Similar to many previous methods, STN aligns both the marginal and conditional distributions across domains. However, different from them, STN adopts the soft-label strategy of unlabeled target data to match the conditional distributions, which averts the hard assignment of each unlabeled target data. Furthermore, an adaptive coefficient is used to gradually increase the importance of the soft-labels. Experiments on three types of transfer tasks testify the effectiveness of STN. As a future direction, we plan to embed adversarial learning strategies into the STN by using, for instance, a domain discriminator. Another interesting direction is to apply the proposed Soft-MMD loss to other fields including homogeneous DA and few-shot learning.

\section*{Acknowledgments}

This research was supported in part by National Key R\&D Program of China under Grant No. 2018YFB0504905, and Shenzhen Science and Technology Program under Grant Nos. JCYJ20180507183823045, JCYJ20160330163900579, and JCYJ20170811160212033, and Natural Science Foundation of China under Grant No. 61673202.

%
\bibliographystyle{ACM-Reference-Format}
\balance
\bibliography{STN}

\end{document}